\documentclass[preprint, 3p]{elsarticle}

\usepackage{lineno,hyperref}
\usepackage[linesnumbered,vlined,ruled]{algorithm2e}
\usepackage{amsfonts}
\usepackage{amsmath}
\usepackage{amssymb}
\usepackage{scalerel}
\usepackage{booktabs}
\usepackage{caption}
\usepackage{subfig}
\usepackage{threeparttable}
\usepackage[figuresright]{rotating}
\usepackage{tikz}

\SetAlFnt{\small}
\modulolinenumbers[5]
\SetAlgoCaptionSeparator{.}

\newdefinition{algo1}{Algorithm}
\newdefinition{rmk}{Remark}

\newenvironment{pcode}[1][htb]
{
\setcounter{algocf}{\value{algo1}}
\begin{algorithm}[#1]%
}{\end{algorithm}
\stepcounter{algo1}
}

\newcommand{\ra}[1]{\renewcommand{\arraystretch}{#1}}

\newcommand{\acol}{p{1.85cm}}
\newcommand{\tcol}{p{1.20cm}}

\usetikzlibrary{shapes.geometric, arrows, chains}
\tikzstyle{process} = [rectangle, minimum width=2cm, minimum height=2cm, text width = 3cm, text  centered, draw=black]
\tikzstyle{para} = [rectangle, minimum width=1cm, minimum height=1cm, text width = 3cm, text  centered]
\tikzstyle{style} = [rectangle, minimum width=2cm, minimum height=2cm, text width = 3cm, text  centered]
\tikzstyle{io} = [trapezium, trapezium left angle=70, trapezium right angle=110, minimum width=1cm, minimum height=1cm, text width = 2cm, text centered, draw=black]
\tikzstyle{arrow} = [thick, ->, >=stealth]

\journal{arXiv}

\begin{document}

\begin{frontmatter}

\title{An enhanced KNN-based twin support vector machine with stable learning rules}

\author[mir]{A. Mir}
\ead{a.mir@iau-tnb.ac.ir}

\author[mysecondaryaddress]{Jalal A. Nasiri\corref{mycorrespondingauthor}}
\cortext[mycorrespondingauthor]{Corresponding author}
\ead{j.nasiri@irandoc.ac.ir}

\address[mir]{Faculty of Electrical and Computer Engineering, North Tehran Branch, Islamic Azad University, Tehran, Iran}
\address[mysecondaryaddress]{Iranian Research Institute for Information Science and Technology (IranDoc), Tehran, Iran}

\begin{abstract}
Among the extensions of twin support vector machine (TSVM), some scholars have utilized K-nearest neighbor (KNN) graph to enhance TSVM’s classification accuracy. However, these KNN-based TSVM classifiers have two major issues such as high computational cost and overfitting. In order to address these issues, this paper presents an enhanced regularized K-nearest neighbor based twin support vector machine (RKNN-TSVM). It has three additional advantages: (1) Weight is given to each sample by considering the distance from its nearest neighbors. This further reduces the effect of noise and outliers on the output model. (2) An extra stabilizer term was added to each objective function. As a result, the learning rules of the proposed method are stable. (3) To reduce the computational cost of finding KNNs for all the samples, location difference of multiple distances based k-nearest neighbors algorithm (LDMDBA) was embedded into the learning process of the proposed method. The extensive experimental results on several synthetic and benchmark datasets show the effectiveness of our proposed RKNN-TSVM in both classification accuracy and computational time. Moreover, the largest speedup in the proposed method reaches to 14 times.
\end{abstract}


\end{frontmatter}


\section{Introduction} \label{sec:1}

Support Vector Machine (SVM) proposed by Vapnik et al.\cite{Cortes1995}, is a state-of-the-art binary classifier. It is on the basis of Statistical Learning Theory (SLT) and Structural Risk Minimization (SRM) \cite{vapnik1999overview}. Due to the SVM's great generalization ability, it has been applied successfully in a wide variety of applications, such as arrhythmia detection \cite{nasiri2009ecg}, Wi-Fi indoor location \cite{figuera2012}, impulse noise removal \cite{roy2016}, and color image watermarking \cite{wang2017}. Its main idea is to find an optimal separating hyperplane between two classes of samples by solving a complex Quadratic Programming Problem (QPP) in the dual space. 

Researchers have proposed many classifiers on the basis of SVM \cite{Nayak2015}. For example, Fung and Mangasrian \cite{Mangasarian} proposed proximal support vector machine (PSVM) which generates two parallel hyperplanes for classifying samples instead of a single hyperplane. In 2002, Lin and Wang proposed \cite{Lin2002} fuzzy support vector machine (FSVM) which introduces fuzzy membership of samples to each of the classes. As a result, the output model of FSVM is less sensitive to noise and outliers. Later Mangasrian and Wild \cite{Mangasarian2006} proposed generalized eigenvalue proximal SVM (GEPSVM) on the basis of PSVM. It generates two non-parallel hyperplanes such that each plane is closest to one of the two classes and as far as possible from the other class. 

In 2007, Jayadeva et al. \cite{Khemchandani2007} proposed twin support vector machine (TSVM) to reduce the computational complexity of standard SVM. TSVM does classification by generating two non-parallel hyperplanes. Each of which is as close as possible to one of the two classes and as far as possible from samples of the other class. To obtain two nonparallel hyperplanes, TSVM solves two smaller-sized QPPs. This makes the learning speed of TSVM classifier four times faster than that of SVM in theory.

Over the past decade, many extensions of TSVM have been proposed \cite{ding2014overview, ding2017twin, huang2018twin}. In 2012, Yi et al. \cite{Ye2012} proposed weighted twin support vector machines with local information (WLTSVM). By finding k-nearest neighbors for all the samples, WLTSVM gives different weight to samples of each class based on the number of its nearest neighbors. This approach is better than TSVM in terms of accuracy and computational complexity. It also considers only one penalty parameter as opposed to two in TSVM. In 2014, Nasiri et al. \cite{Nasiri2014} proposed an energy-based least squares twin support vector machine (ELS-TSVM) by introducing an energy parameter for each hyperplane. In ELS-TSVM, different energy parameters are selected according to prior knowledge to reduce the effect of noise and outliers.

In 2015, Pan et al. \cite{Pan2015} proposed K-nearest neighbor based structural twin support vector machine (KNN-STSVM). Similar to S-TSVM \cite{qi2013structural}, this method incorporates the data distribution information by using Ward's linkage clustering algorithm. However, the KNN method applied in S-TSVM to give different weight to each sample and remove redundant constraints. As a result, the classification accuracy and computational complexity of S-TSVM were improved.

In 2016, Xu \cite{xu2016k} proposed K-nearest neighbor-based weighted multi-class twin support vector machine (KWM-TSVM). It embodies inter and intra-class information into the objective function of Twin-KSVC \cite{xu2013twin}. As a result, the computational cost and prediction accuracy of the classifier were improved. Recently, Xu \cite{pang2018scaling} proposed a safe instance reduction to reduce the computational complexity of KWMTSVM. This method is safe and deletes a large portion of samples of two classes. Therefore, the computational cost will be decreased significantly.

It should be noted that many weighted TSVM methods were proposed over the past few years. However, this paper is concerned with KNN-based TSVM methods \cite{Ye2012, Pan2015, xu2016k}. Therefore, it addresses the drawbacks of these methods which are explained as follows:
\begin{enumerate}
	\item These methods give weight to samples of each class solely by counting the number of k-nearest neighbors of each sample. However, they do not consider the distance between pairs of nearest neighbors. To further improve the identification of highly dense samples, weight can be given to a sample with respect to the distance from its nearest neighbors. In other words, a sample with closer neighbors is given higher weight than the one with farther neighbors.
	\item Similar to TSVM, these classifiers minimize the empirical risk in their objective functions, which may lead to the overfitting problem and reduces the prediction accuracy \cite{shalev2014}. To address this issue, the tradeoff between overfitting and generalization can be determined by adding a stabilizer term to each objective function.
	\item These KNN-based classifiers utilize full search algorithm (FSA) to find k-nearest neighbors of each sample. The FSA method has a time complexity of $\mathcal{O}(n^2)$ which is time-consuming for large-scale datasets. However, scholars have proposed new KNN methods which have lower computational cost than that of FSA algorithm. For instance, Xia et al. \cite{xia2015location} proposed location difference of multiple distances based k-nearest neighbors algorithm (LDMDBA). This method can be used to reduce the overall computational complexity of KNN-based TSVM classifiers.
\end{enumerate}

Motivated by the above discussion and studies, we propose an enhanced regularized K-nearest neighbor based twin support vector machine (RKNN-TSVM). Different from other KNN-based TSVM methods \cite{Ye2012, Pan2015, xu2016k}, the proposed method gives weight to each sample with respect to the distance from its nearest neighbors. This further enhances the identification of highly dense samples, outliers and, noisy samples. Moreover, due to the minimization of the SRM principle,  the optimization problems of the proposed method are positive definite and stable.

The high computational cost is the main challenge of our proposed method, especially for large-scale datasets. So far, many fast KNN algorithms were proposed to accelerate finding K-nearest neighbors of samples, including k-dimensional tree (k-d tree) \cite{friedman1977}, a lower bound tree (LB tree) \cite{chen2007fast}, LDMDBA algorithm \cite{xia2015location} and so on. The recently proposed LDMDBA method has a time complexity of $\mathcal{O}(\log{d}n\log{n})$ which is less than the FSA algorithm and most of other KNN methods. In addition, this method does not rely on any tree structure so that it is efficient for datasets of high dimensionality. In this paper, LDMDBA algorithm is introduced into our proposed method to speed up KNN finding.

The main advantages of our proposed method can be summarized as follows:
\begin{itemize}
	\item In comparison with other KNN-based TSVM classifiers \cite{Ye2012, Pan2015, xu2016k}, the proposed method gives weight to samples differently. The weight of each sample was calculated based on the distance between its nearest neighbors. This further improves fitting hyperplanes with highly dense samples. In the proposed method, samples with closer neighbors are weighted more heavily than the one with farther neighbors.
	
	\item The proposed method has two additional parameters for determining the tradeoff between overfitting and generalization. As a result, the learning rules of our RKNN-TSVM are stable and do not overfit the output model to all the training samples.
	
	\item As previously stated, KNN finding reduces significantly the learning speed of our classifier. The LDMDBA algorithm \cite{xia2015location} was employed to further reduce the overall computational complexity of the proposed method. This KNN algorithm has lower time complexity than FSA algorithm. Moreover, LDMDBA algorithm is effective for non-linear case where samples are mapped from input space to higher dimensional feature space.
	
	\item Due to the giving weight to samples w.r.t the distance from their nearest neighbors, the proposed method gives much less weight to noisy samples and outliers. Consequently, the output model is less sensitive and potentially more robust to the outliers and noise.
\end{itemize}

The rest of this paper is organized as follows. Section \ref{sec:2} presents the notation used in the rest of the paper, briefly reviews TSVM, WLTSVM, and LDMDBA algorithm. Section \ref{sec:7} gives the detail of the proposed method, including linear and nonlinear cases. Algorithm analysis of RKNN-TSVM is given in Section \ref{sec:11}. Section \ref{sec:14} discusses the experimental results on synthetic and benchmark datasets to investigate the validity and effectiveness of our proposed method. Finally, the concluding remarks are given in Section \ref{sec:26}.

\section{Backgrounds} \label{sec:2}

This section defines the notation that will be used in the rest of the paper and includes the brief description of conventional TSVM, WLTSVM, and LDMDBA algorithm.

\subsection{Notation}\label{sec:3}
Let $T=\{(x_1, y_1), ... , (x_n, y_n)\}$ be the full training set of $n$ $d$-dimensional samples. where $x_{i} \in \mathbb{R}^{d}$ is a feature vector and $y_{i} \in \{-1, 1\}$ are corresponding labels. Let $X^{(i)} = [x_{1}^{(i)},x_{2}^{(i)},...,x_{n_{i}}^{(i)}],i=1,2$ be a matrix consisting of $n_{i}$ samples that are $d$ dimensional in class $i$, $X^{(i)} \in T$, $X^{(i)} \in \mathbb{R}^{n_{i} \times d}$. For convenience, matrix $A$ in $\mathbb{R}^{n_1 \times d}$ represents the samples of class $1$ and matrix $B$ in $\mathbb{R}^{n_2 \times d}$ represents the samples of class $-1$, where $n_1 + n_2 = n$. Table \ref{tab:1} provides a summary of the notation used in this paper.

\begin{table}[t]
	\small
	\centering
	\caption{Summary of notation used throughout the paper.}
	\begin{tabular}{l l}
		\toprule
		Definition & Notation \\
		\midrule
		Number of samples & $n$ \\
		Number of input features & $d$\\
		Sample $i$ & $x_{i} \in \mathbb{R}^{d}$ \\
		Label of sample $i$ & $y_{i} \in \{-1, 1\}$\\
		Full training set & $T=\{(x_1, y_1),\dots,(x_n, y_n)\}$\\
		Samples of class $+1$ and $-1$ & $A \in \mathbb{R}^{n_1 \times d}$, $B \in \mathbb{R}^{n_2 \times d}$ \\
		Column vectors of ones & $e_{1} \in \mathbb{R}^{n_{1} \times 1}$, $e_{2} \in \mathbb{R}^{n_{2} \times 1}$ \\
		Identity matrix & $I$ \\
		Slack vectors & $\xi$, $\eta$\\
		Lagrangian multipliers & $\alpha \in \mathbb{R}^{n_{2}} $, $\beta \in \mathbb{R}^{n_{1}}$ \\
		Norm & $\left\|\ .\ \right\|:\mathbb{R}^{d}\mapsto\mathbb{R}$ \\
		Weights of hyperplane $i$ & $w_{i} \in \mathbb{R}^{d}(i=1,2)$ \\
		Bias of hyperplane $i$ & $b_{i} \in \mathbb{R}(i=1,2)$\\
		\bottomrule
	\end{tabular}
	
	\label{tab:1}
\end{table}

\subsection{Twin support vector machine} \label{sec:4}
TSVM \cite{Khemchandani2007} is binary classifier whose idea is to find two non-parallel hyperplanes. To explain this classifier, consider a binary classification problem of $n_{1}$ samples belonging to class $+1$ and $n_{2}$ samples belonging to class $-1$ in the $d$-dimensional real space $\mathbb{R}^{d}$. The linear TSVM \cite{Khemchandani2007} seeks a pair of non-parallel hyperplanes as follows:
\begin{equation}\label{eq:1}
{{x}^T}{{w}_{1}}+{{b}_{1}}=0 \quad \textrm{and} \quad {{x}^T}{{w}_{2}}+{{b}_{2}}=0
\end{equation}
\noindent such that each hyperplane is closest to the samples of one class and far from the samples of other class, where $w_{1} \in \mathbb{R}^{d}$, $w_{2} \in \mathbb{R}^{d}$,$b_{1} \in \mathbb{R}$ and $b_{2} \in \mathbb{R}$.

To obtain the above hyperplanes (\ref{eq:1}), TSVM solves two primal QPPs with objective function corresponding to one class and constrains corresponding to other class.

\begin{equation}\label{eq:2}
\begin{split}
\mathop{{ min}}\limits_{w_{1} ,b_{1}} \qquad & \frac{1}{2}{{\left\| A{{w}_{1}}+{{e}_{1}}{{b}_{1}} \right\|}^{2}}+{{c}_{1}}e_{2}^{T}\xi \\
\textrm{s.t. } \qquad & -(B{{w}_{1}}+{{e}_{2}}{{b}_{1}})+\xi\ge {{e}_{2}}\text{ },\xi\ge 0
\end{split}
\end{equation}
\begin{equation}\label{eq:3}
\begin{split}
\mathop{{ min}}\limits_{w_{2} ,b_{2}} \qquad & \frac{1}{2}{{\left\| B{{w}_{2}}+{{e}_{2}}{{b}_{2}} \right\|}^{2}}+{{c}_{2}}e_{1}^{T}\eta \\
\textrm{s.t. } \qquad & (A{{w}_{2}}+{{e}_{1}}{{b}_{2}})+\eta\ge {{e}_{1}}\text{ },\eta\ge 0
\end{split}
\end{equation}

\noindent where $c_1$ and $c_2$ are positive penalty parameters, $\xi_{1}$ and $\xi_{2}$ are slack vectors, $e_1$ is the column vectors of ones of $n_{1}$ dimensions and $e_{2}$ is the column vectors of ones of $n_{2}$ dimensions.

By introducing Lagrangian multipliers $\alpha \in \mathbb{R}^{n_{2}} $ and $\beta \in \mathbb{R}^{n_{1}}$, the Wolfe dual of QPPs (\ref{eq:2}) and (\ref{eq:3}) are given by:

\begin{equation}\label{eq:4}
\begin{split}
\mathop{{ min}}\limits_{\alpha} \qquad & \frac{1}{2}{{\alpha }^{T}}G{{({{H}^{T}}H)}^{-1}}{{G}^{T}}\alpha -e_{2}^{T}\alpha  \\
\textrm{s.t. } \qquad & 0{{e}_{2}}\le \alpha \le {{c}_{1}}{{e}_{2}}
\end{split}
\end{equation}
\begin{equation}\label{eq:5}
\begin{split}
\mathop{{ min}}\limits_{\beta} \qquad & \frac{1}{2}{{\beta }^{T}}H{{({{G}^{T}}G)}^{-1}}{{H}^{T}}\beta -e_{1}^{T}\beta  \\
\textrm{s.t. } \qquad & 0{{e}_{1}}\le \beta \le {{c}_{2}}{{e}_{1}}
\end{split}
\end{equation}

\noindent where $H=[A\text{ }e]$ and $G=[B\text{ }e]$. From the dual problems of (\ref{eq:4}) and (\ref{eq:5}), one can notice that QPPs (\ref{eq:4}) and (\ref{eq:5}) have $n_1$ and $n_2$ parameters, respectively, as opposed to $n=n_{1}+n_{2}$ parameters in standard SVM.

After solving the dual QPPs (\ref{eq:4}) and (\ref{eq:5}), the two non-parallel hyperplanes are given by:

\begin{align}
\left[ \begin{aligned}
& {{w}_{1}} \\
& {{b}_{1}} \\
\end{aligned}\right]&= -{{({{H}^{T}}H)}^{-1}}{{G}^{T}}\alpha \label{eq:6} \\
\left[ \begin{aligned}
& {{w}_{2}} \\
& {{b}_{2}} \\
\end{aligned}\right]&= {{({{G}^{T}}G)}^{-1}}{{H}^{T}}\beta \label{eq:7}
\end{align}

In addition to solving dual QPPs (\ref{eq:4}) and (\ref{eq:5}), TSVM also requires inversion of matrices ${{H}^{T}}H$ and ${{G}^{T}}G$ which are of size $(d+1)\times(d+1)$ where $d \ll n$.

A new testing sample $x \in \mathbb{R}^{d}$ is assigned to class $i(i=-1,+1)$ by
\begin{equation} \label{eq:8}
\underset{{}}{\mathop{Class\text{ }i\text{ }=\underset{j=1,2}{\mathop{arg\min }}\,}}\,\text{ }\frac{\left| {{x}^{T}}{{w}_{j}}+{{b}_{j}} \right|}{\left\|w_j\right\|}
\end{equation}
where $|.|$ denotes the perpendicular distance of sample $x$ from the hyperplane. TSVM was also extended to handle non-linear kernels by using two non-parallel kernel generated-surfaces \cite{Khemchandani2007}.

In TSVM, if the number of samples in two classes is approximately equal to $n/2$, then its computational complexity is $\mathcal{O}(1/4n^{3})$. This implies that TSVM is approximately four times faster than standard SVM in theory \cite{Khemchandani2007}.

\subsection{Weighted twin support vector machine with local information}
\label{sec:5}
\begin{figure}[!t]
	\centering
	\includegraphics[width=0.5\columnwidth]{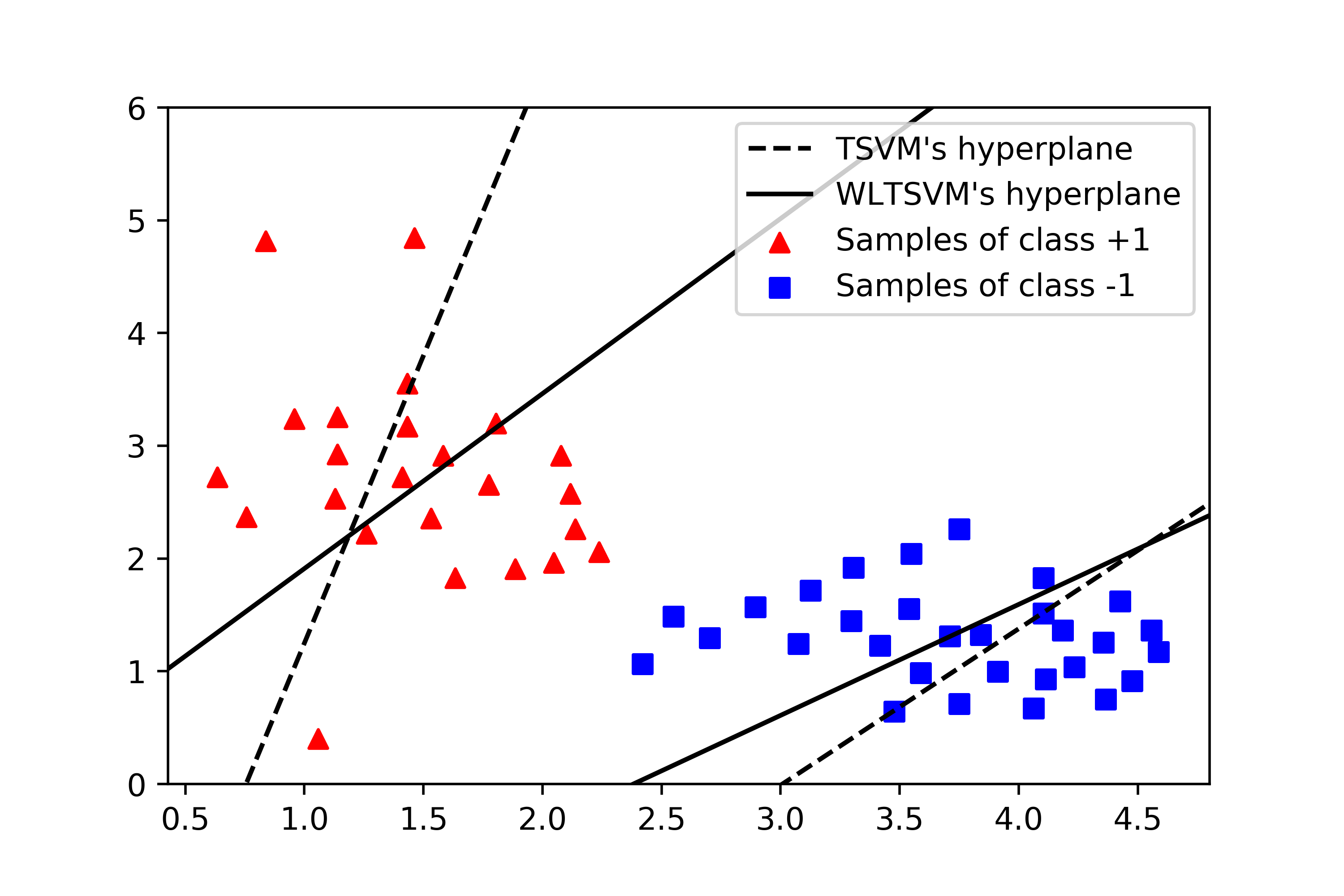}
	\caption{The geometric comparison of standard TSVM with WLTSVM classifier.}
	\label{fig:1}
\end{figure}

One of the issues of TSVM is that it fails to determine the contribution of each training sample to the output model. Therefore, its output model becomes sensitive to noise and outliers. WLTSVM \cite{Ye2012} addressed this issue by finding the KNNs of all training samples. This method constructs intra-class graph $W_{s}$ and inter-class graph $W_{d}$ to embed weight of each sample into optimization problems of TSVM. As a result, it fits samples with high-density as opposed to TSVM whose hyperplane fits all the samples of its own class. Fig. \ref{fig:1} indicates a geometrical comparison between linear TSVM and linear WLTSVM classifier in two-dimensional real space $\mathbb{R}^{2}$. As shown in Fig. \ref{fig:1}, WLTSVM is less sensitive to outliers and noisy samples than TSVM.

WLTSVM solves a pair of smaller sized QPPs as follows:

\begin{equation}\label{eq:9}
\begin{split}
\mathop{{ min}}\limits_{w_{1} ,b_{1}} \qquad & \frac{1}{2}\sum\limits_{i=1}^{n_{1}}{\sum\limits_{j=1}^{n_{1}}{W^{(1)}_{s,ij}(w^{T}_{1}x^{(1)}_{j}+b_{1})^{2}}}+c\sum\limits_{j=1}^{n_{2}}\xi_j \\
\textrm{s.t. } \qquad & -f^{(2)}_{j}(w^{T}_{1}x^{(2)}_{j}+b_{1})+\xi_{j} \ge f^{(2)}_{j} \\
& \xi_{j} \ge 0,\quad j=1,...,n_{2}
\end{split}
\end{equation}
\begin{equation}\label{eq:10}
\begin{split}
\mathop{{ min}}\limits_{w_{2} ,b_{2}} \qquad & \frac{1}{2}\sum\limits_{i=1}^{n_{2}}{\sum\limits_{j=1}^{n_{2}}{W^{(2)}_{s,ij}(w^{T}_{2}x^{(2)}_{j}+b_{2})^{2}}}+c\sum\limits_{j=1}^{n_{1}}\eta_j \\
\textrm{s.t. } \qquad & f^{(1)}_{j}(w^{T}_{2}x^{(1)}_{j}+b_{2})+\eta_{j} \ge f^{(1)}_{j} \\
& \eta_{j} \ge 0,\quad j=1,...,n_{2}
\end{split}
\end{equation}

\noindent In the optimization problems of WLTSVM (\ref{eq:9}) and (\ref{eq:10}), different weights are given to the samples of each class according to their KNNs. Unlike TSVM, the optimal hyperplane should be far from the margin points instead of all the samples of other class. This further reduces the time complexity by keeping only margin points in the constraints. Moreover, WLTSVM has only one penalty parameter as opposed to two in TSVM.

To obtain the solution, the dual problems for (\ref{eq:9}) and (\ref{eq:10}) are solved, respectively:

\begin{equation}\label{eq:11}
\begin{split}
\mathop{{ min}}\limits_{\alpha} \qquad & \frac{1}{2}{{\alpha }^{T}}(F^{T}G){{({{H}^{T}}DH)}^{-1}}{({G}^{T}F)}\alpha -e_{2}^{T}F\alpha  \\
\textrm{s.t. } \qquad & 0{{e}_{2}}\le \alpha \le {{c}}{{e}_{2}}
\end{split}
\end{equation}
\begin{equation}\label{eq:12}
\begin{split}
\mathop{{ min}}\limits_{\beta} \qquad & \frac{1}{2}{{\beta }^{T}}(P^{T}H){{({{G}^{T}}QG)}^{-1}}{({H}^{T}P)}\beta -e_{1}^{T}P\beta  \\
\textrm{s.t. } \qquad & 0{{e}_{1}}\le \beta \le {{c}}{{e}_{1}}
\end{split}
\end{equation}

\noindent where $D=diag(d^{(1)}_{1},d^{(1)}_{2},\dots,d^{(1)}_{n_{1}})$, $Q=diag(d^{(2)}_{1},d^{(2)}_{2}, \allowbreak \dots,d^{(2)}_{n_{2}})$, $F=diag(f^{(2)}_{1},f^{(2)}_{2},\dots,f^{(2)}_{n_{2}})$ and $P=diag(f^{(1)}_{1}, \allowbreak f^{(1)}_{2},\dots,f^{(1)}_{n_{1}})$ are diagonal matrices, respectively ($f_{j}$ is either 0 or 1.). Both $e_1$ and $e_2$ are vectors of all ones of $n_1$  and $n_2$ dimensions, respectively.

Similar to TSVM, a new sample is classified as class $+1$ or class $-1$ depends on which of the two hyperplanes it lies nearest to. Although WLTSVM has clear advantages over TSVM such as better classification ability and less computational cost, it has the following drawbacks:
\begin{enumerate}
	\item WLTSVM gives different treatments and weight to each sample by only counting the number of its nearest neighbors. For instance, the weight of each sample in class $+1$ can be computed as follows:
	\begin{equation}\label{eq:13}
	{{d}_{j}^{(1)}}=~\underset{i=1}{\overset{{{n}_{1}}}{\mathop \sum }}\,{{W}_{s,ij}}~,~j=1,2,\ldots ,{{n}_{1}}
	\end{equation}
	\noindent where $d_{j}^{(1)}$ denotes the weight of sample $x_{j}$. It should be noted that $W_{s,ij}$ is either 0 or 1. This implies that WLTSVM treats the nearest neighbors of each sample similarly. Therefore, the weight matrix $W_{s,ij}$ contains only binary values. 
	\item  In order to deal with matrix singularity, the inverse matrices $({H}^{T}DH)^{-1}$ and $({G}^{T}QG)^{-1}$ are approximately replaced by $({H}^{T}DH + \varepsilon I)^{-1}$ and $({G}^{T}QG + \varepsilon I)^{-1}$, respectively, where $\varepsilon$ is a positive scalar. Hence only approximate solutions to (\ref{eq:11}) and (\ref{eq:12}) are obtained.
	
	\item Although WLTSVM reduces the time complexity by keeping only margin points in the constraints, it has to find k-nearest neighbors for all the samples. Consequently, the overall computational complexity of WLTSVM is about $\mathcal{O}(2n_{1}^{3}+n^{2}logn)$ under the assumption that $n_{1}=n_{2}$, where $n_{1},n_{2} \ll n$. This makes WLTSVM impractical for large-scale datasets. To mitigate this problem, fast KNN methods can be utilized.
\end{enumerate}
The proposed method addresses these issues.

\subsection{Location difference of multiple distances based nearest neighbors searching algorithm (LDMDBA)} \label{sec:6}
The LDMDBA algorithm \cite{xia2015location} introduced the concept of location difference among different samples. The central idea of this method is that the nearest neighbors of each sample can be found when their distance from some reference points is known. Due to this idea, LDMDBA algorithm avoids computing distance between each pair of samples.

Consider the KNN finding problem with training set $T$ (defined in Table \ref{tab:1}), a sample $x_{j} \in T$ and the distance from reference point $O_{1}$ to $x_j$ is denoted as $Dis_{1}(x_{j})=\left\|x_{j}-O_{1}\right\|$. According to \cite{xia2015location}, the number of reference points is taken as $\log_{2}{d}$. The values of the first $i$ dimensions of the $i$th reference point $O_{i}$ can be set to $-1$ and other values are set to $1$ (i.e. $O_{i}=(-1,-1,\dots,-1,1,\dots,1)$, where the number of values $-1$ is equal to $i$). The neighbors of the sample $x_j$ found using the $i$th reference point are denoted by $Nea_{i}(x_{j})$.

To compute $Nea_{i}(x_{j})$, the distance from all the reference points to the sample $x_{i}$ are first computed. After sorting the distance values, a sorted sequence is obtained. The $k$-nearest neighbors of sample $x_{j}$ are mostly located in a subsequence with the center sample $x_{j}$ in the sequence. The length of the subsequence can be denoted as $2k*\varepsilon$ where $\varepsilon$ is set to $\log_{2}{\log_{2}{n}}$ (more information on how the value $\varepsilon$ was determined can be found in \cite{xia2015location}). Finally, all the exact Euclidean distance between samples in the subsequence are computed. Those samples corresponding to the $k$-smallest distances in the subsequence can be considered as the $k$-nearest neighbor of the sample $x_{j}$. For clarity, the LDMDBA algorithm is explicitly stated.

\begin{algo1}
	LDMDBA (Location Difference of Multiple Distances-based Algorithm)
	
	Given a training set $T$, let $k$ be the number of nearest neighbors in the algorithm. Starting with $i=1$, the $k$-nearest neighbors of each sample $x_{j} \in T$ can be obtained using the following steps:
	
	\begin{enumerate}
		\item The $i$th reference point $O_{i}$ is set as a vector whose values of the first $i$ dimensions are equal to $-1$, and the other values are set to $1$.
		\item Compute the distance from $i$th reference point $O_{i}$ to all the samples using $Dis_{i}(x_{j})=\left\|x_{j}-O_{i}\right\|, \forall i \in \{1,\dots,\log_{2}{d}\}$.
		\item Sort the samples by the values of $Dis_{i}$ and generate a sorted sequence.
		\item For a subsequence of the samples with the fixed range $2k*\log_{2}{\log_{2}{n}}$ and center sample $x_{j}$, compute all the exact Euclidean distances from the sample $x_{j}$ to the samples in the subsequence.
		\item Sort the distance values obtained in the step 4.
		\item The $k$-smallest Euclidean distances in the sorted subsequence are $k$-nearest neighbors of the sample $x_j$.
		\item If the neighbors of all the samples using all the reference points have been computed, terminate;otherwise set $i=i+1$, and go to the step 1.
	\end{enumerate}
\end{algo1}

In Algorithm 1, the time complexity of step 3 and step 5 is determined by the used sorting algorithm which is $\mathcal{O}(n\log_{2}{n})$. Therefore, the overall computational complexity of the LDMDBA algorithm is $\mathcal{O}(\log{d}n\log{n})$. However, the FSA algorithm has a time complexity of $\mathcal{O}(n^{2}\log_{2}n)$ as described in the Algorithm \ref{algo_fsa}.

Moreover, LDMDBA algorithm does not rely on any dimensionality dependent tree structure. As a result, it can be effectively applied to various high dimensional datasets. The experimental results of \cite{xia2015location} indicate the effectiveness of LDMDBA algorithm over FSA and other existing KNN algorithms.

\begin{pcode}[t]
	\SetKwInOut{Input}{input}
	\SetKwInOut{Output}{output}
	\SetKwData{kp}{k}
	\SetKwArray{iKNN}{idxKNN}
	\SetKwArray{ds}{DS}
	\SetKwArray{Kw}{distMat}
	\SetKwArray{tknn}{tempKNN}
	\SetKwArray{tidx}{tempIdx}
	\SetKwFunction{argS}{argSort}
	
	\Input{$T$ : Full training set \\ $k$ : Number of nearest neighbors}
	\BlankLine
	\Output{\iKNN : Indices of KNNs for every sample}
	\BlankLine
	
	\emph{Step 1: Compute the Euclidean distances}\;
	\Kw \emph{: A matrix of size $n \times n$ that holds distances}\;
	\For{$x_{i} \in T$}{
		\For{$x_{j} \in T$}{
			
			\eIf{$i \neq j$}{
				\eIf{$j > i$}{
					\Kw{i, j} $\leftarrow \sqrt{(x_{j} - x^{i}_{j})^{T}(x_{j} - x^{i}_{j})}$\;
					\BlankLine
					
				}(\tcp*[h]{Distance already computed.}){
					\Kw{i, j} $\leftarrow$ \Kw{j, i}\;
				}
				
			}(\tcp*[h]{Distance of i-th point from itself.}){
				\Kw{i, j} $\leftarrow 0$\;
				
			}
		}
	}
	
	\emph{Step 2: Find k-nearest neighbors}\;
	\For{$i \leftarrow 1$ \KwTo $n$}{
		
		\tcp{Indices of nearest neighbors of ith sample.}
		\tidx $\leftarrow$ \argS{\Kw{i, :}}\;
		
		\tcp{K-nearest-neighbors of i-th sample.}
		\tknn $\leftarrow$ \tidx{2:\kp}\;
		
		\For{$l \leftarrow 1$ \KwTo \kp}{
			\iKNN{i, l} $\leftarrow$  \tknn{l}\;
		}
		
	}
	\caption{Full search algorithm (FSA)}\label{algo_fsa}
\end{pcode}

\section{Regularized k-nearest neighbor based twin support vector machine (RKNN-TSVM)} \label{sec:7}
In this section, we present our classifier called regularized k-nearest neighbor based twin support vector machine (RKNN-TSVM). It gives weight to each sample with respect to the distance from its nearest neighbors. Also, the proposed method avoids overfitting by considering the SRM principle in each objective function. 

\subsection{The definition of weight matrices} \label{sec:8}
As discussed in Section \ref{sec:1}, the existing KNN-based TSVM classifiers \cite{Ye2012, Pan2015, xu2016k} constructs a $k$-nearest neighbor graph $G$ to exploit similarity among samples. In these methods, the weight of $G$ is defined as: 
\begin{equation}\label{eq:14}
W_{ij} =
\begin{cases}
1, & \mbox{if $x_i \in Nea\left(x_j\right)$ or $x_j \in Nea\left(x_i\right)$},  \\
0, & \mbox{otherwise}.
\end{cases}
\end{equation}

\noindent where $Nea(x_{j})$ stands for the set of $k$-nearest neighbors of the sample $x_{j}$ which is defined as:
\begin{equation}\label{eq:15}
Nea(x_j) = \{x^{i}_{j} \mid \textrm{if } x^{i}_{j} \textrm{ is a knn of } x_{j} ,1 \leq i \leq k \}
\end{equation}
\noindent the set $Nea(x_{j})$ is arranged in an increasing order in terms of Euclidean distance $d(x_{j},x^{i}_{j})$ between $x_{j}$ and $x^{i}_{j}$.
\begin{equation}\label{eq:16}
d(x_{j},x^{i}_{j})= \sqrt{(x_{j} - x^{i}_{j})^{T}(x_{j} - x^{i}_{j})}
\end{equation}

However, the value of $W_{ij}$ is either $0$ or $1$. This implies that weight of the sample $x_{j}$ is obtained by solely counting the number of its nearest neighbors. To address this issue, weight can be given to a sample based on the distance between its nearest neighbors. Motivated by \cite{dudani1976distance, gou2012new}, the matrix of $G$ is redefined as follows:
\begin{equation}\label{eq:17}
W_{ij} =
\begin{cases}
\acute{w_{ij}}, & \textrm{if } x_{i} \in Nea\left(x_{j}\right) \textrm{ or } x_j \in Nea\left(x_i\right), \\
0, & \textrm{otherwise}.
\end{cases}
\end{equation}
\noindent where $\acute{w_{ij}}$ is the weight of $i$-th nearest neighbor of the sample $x_{j}$ which is given by:
\begin{equation}\label{eq:18}
\acute{w_{ij}} =
\begin{cases}
\frac{d(x_{i},x^{k}_{j}) - d(x_{i},x_{j})}{d(x_{i},x^{k}_{j}) - d(x_{i},x^{1}_{j})}, & \textrm{if } d\left(x_{i},x^{k}_{j} \right) \neq d\left(x_{i},x^{1}_{j} \right),  \\
1, & \textrm{if } d\left(x_{i},x^{k}_{j} \right) = d\left(x_{i},x^{1}_{j} \right).
\end{cases}
\end{equation}

According to the Eq. (\ref{eq:18}), It can be noted that a neighbor $x_{i}$ with smaller distance is weighted more heavily than the one with the greater distance. Therefore, the values of $\acute{w_{ij}}$ are scaled linearly to the interval $\left[0,1\right]$.

Similar to (\ref{eq:17}), the weight matrices for class $+1$ and $-1$ are defined in (\ref{eq:19}) and (\ref{eq:20}), respectively.
\begin{equation}\label{eq:19}
W_{s,ij} =
\begin{cases}
\acute{w_{ij}}, & \textrm{if } x_{i} \in Nea_{s}\left(x_{j}\right) \textrm{ or } x_j \in Nea_{s}\left(x_i\right), \\
0, & \textrm{otherwise}.
\end{cases}
\end{equation}
\begin{equation}\label{eq:20}
W_{d,ij} =
\begin{cases}
\acute{w_{ij}}, & \textrm{if } x_{i} \in Nea_{d}\left(x_{j}\right), \\
0, & \textrm{otherwise}.
\end{cases}
\end{equation}

\noindent where $Nea_{s}(x_{j})$ stands for the $k$-nearest neighbors of the sample $x_{j}$ in the class $+1$ and $Nea_{d}(x_{j})$ denotes the $k$-nearest neighbors of the sample $x_{j}$ in the class $-1$. Specifically,
\begin{align}
Nea_{s}\left(x_j\right) = \{x^{i}_{j} \mid l(x^{i}_{j}) = l(x_{j}), 1 \leq i \leq k \} \label{eq:21} \\
Nea_{d}\left(x_j\right) = \{x^{i}_{j} \mid l(x^{i}_{j}) \neq l(x_{j}), 1 \leq i \leq k \}
\end{align}

\noindent where $l(x_{j})$ denotes the class label of the sample $x_{j}$. Clearly, $Nea_{s}(x_j)\,\cap\, Nea_{d}(x_j)= \varnothing$ and $Nea_{s}(x_j)\, \cup \, Nea_{d}(x_j) = Nea(x_j)$. When $W_{s,ij} \neq 0$ or $W_{d,ij} \neq 0$, an undirected edge between node $x_{i}$ and $x_{j}$ is added the the corresponding graph.

Unlike TSVM, only the support vectors (SVs) instead of all the samples of the other class are important for optimal production of the hyperplane of the corresponding class. To directly extract possible SVs (margin points)  from the samples in class $-1$, we redefine the weight matrix $W_{d}$ as follows:
\begin{equation}\label{eq:23}
f_{j} =
\begin{cases}
1, & \exists j, W_{d,ij} \neq 0,  \\
0, & \textrm{otherwise}.
\end{cases}
\end{equation}

The procedure of computing weight of samples and extracting margin points are outlined in the Algorithm \ref{algo_weight_margin}.

\begin{pcode}[!t]
	
	\SetKw{Kwand}{and}
	\SetKwInOut{Input}{input}
	\SetKwInOut{Output}{output}
	\SetKwArray{iKNN}{idxKNN}
	\SetKwArray{ic1}{idxC1}
	\SetKwFunction{cnz}{countNonZero}

	\Input{$X^{(i)} = [x_{1}^{(i)},x_{2}^{(i)},...,x_{n_{i}}^{(i)}],i=1,2$ \\ \iKNN : Indices of KNNs for every sample}
	\BlankLine
	\Output{$d^{(1)}_1, \dots, d^{(1)}_{n_{1}}$ : Weight of samples in class $+1$ \\ $f^{(2)}_{1},\dots,f^{(2)}_{n_{2}}$ : Margin points of class $-1$}
	\BlankLine
	
	%
	%
	%
	%
	%
	%
	
	$W_{s}$ \tcp{Matrix of size $n_{1} \times n_{1}$ for within-class graph}
	
	\For{$x_{i} \in X^{(1)}$}{
		
		\For{$x_{j} \in X^{(1)}$}{
			
			\uIf{$(i \neq j)$ \Kwand $(x_{i} \in Nea_{s}(x_{j}))$}{
				
				\uIf{$d\left(x_{i},x^{k}_{j} \right) \neq d\left(x_{i},x^{1}_{j} \right)$}{
					
					$W_{s,ij} \leftarrow  \frac{d(x_{i},x^{k}_{j}) - d(x_{i},x_{j})}{d(x_{i},x^{k}_{j}) - d(x_{i},x^{1}_{j})}$ \;
					
				}
				\Else{
					$W_{s,ij} \leftarrow 1 $\;
				}
				
			}
			\uElseIf{$i = j$}{
				
				$W_{s,ij} \leftarrow 1$\;
				
			}
			\Else{
				$W_{s,ij} \leftarrow 0$\;
			}	
		}
	}

	$W_{d}$ \tcp{Matrix of size $n_{1} \times n_{2}$ for between-class graph}
	
	\For{$x_{i} \in X^{(1)}$}{
		
		\For{$x_{j} \in X^{(2)}$}{
			
			\eIf{$x_{i} \in Nea_{d}(x_{j})$}{
				
				$W_{d,ij} \leftarrow \frac{d(x_{i},x^{k}_{j}) - d(x_{i},x_{j})}{d(x_{i},x^{k}_{j}) - d(x_{i},x^{1}_{j})}$ \;
				
			}{
				
				$W_{d,ij} \leftarrow 0$\;			
				
			}
		}
	}
	
	\For{$j \leftarrow 1$ \KwTo $n_{1}$}{
		
		$d_{j} \leftarrow \sum^{n_1}_{i=1} W^{(1)}_{s,ij}$\;
		
	}
	\For{$j \leftarrow 1$ \KwTo $n_{2}$}{
		
		\eIf{ $\exists j, W_{d,ij} \neq 0$ }{
			
			$f_j \leftarrow 1$\;
		}{
			
			$f_j \leftarrow 0$\;
		}		
	}
	
	%
	%
	%
	%
	%
	
	\caption{The computation of weight matrices}\label{algo_weight_margin}
\end{pcode}

\subsection{Linear case} \label{sec:9}
\begin{figure}[!t]
	\centering
	\includegraphics[width=0.5\columnwidth]{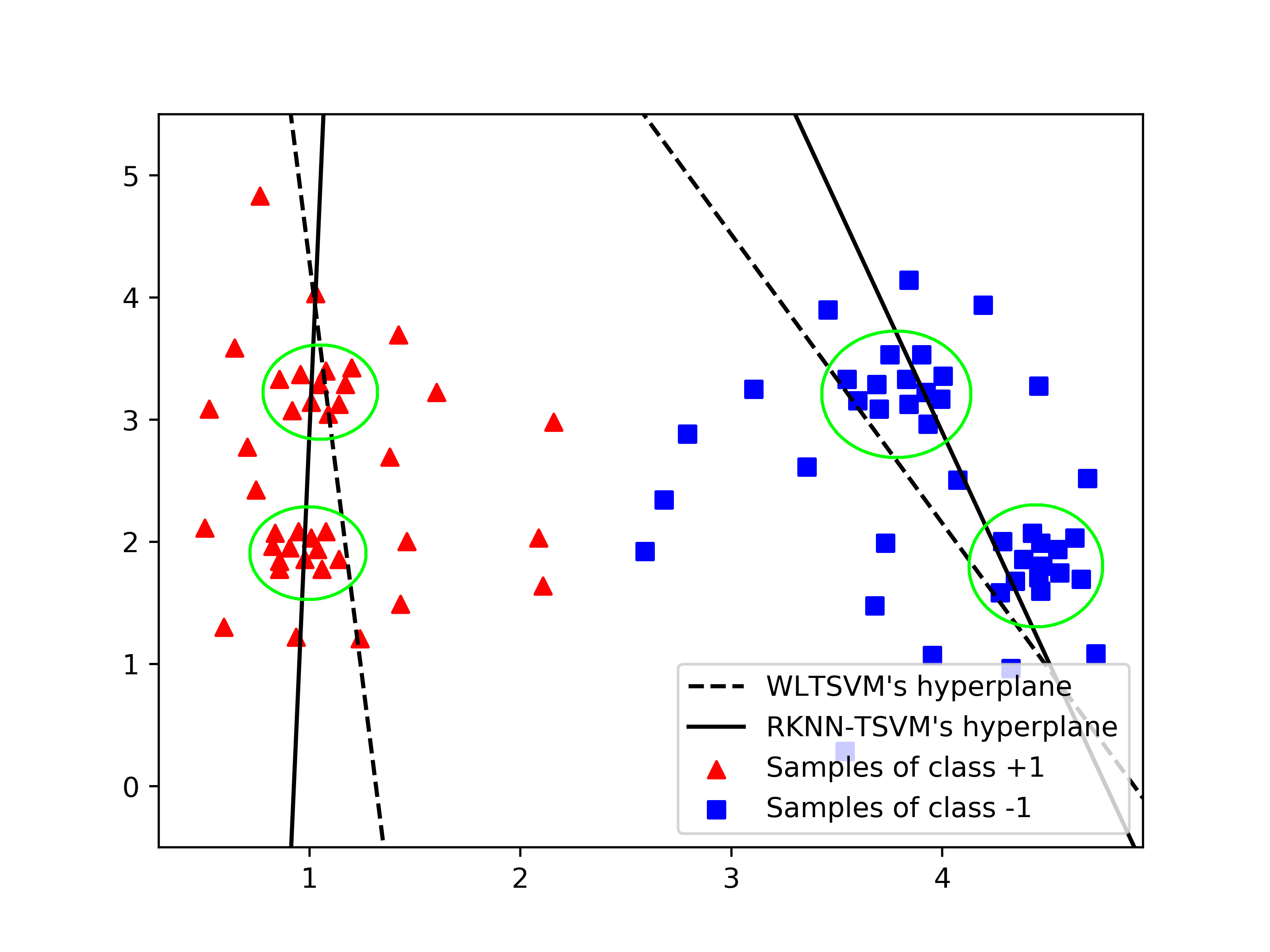}
	\caption{The basic thought of our RKNN-TSVM classifier. The high-density samples are denoted by green circles.}
	\label{fig:4}
\end{figure}

As stated in section \ref{sec:8}, the distance of a sample from its nearest neighbors plays an important role in finding highly dense samples. Following this, the yielded hyperplane is closer to highly dense samples of its own class.  Fig. \ref{fig:4} shows the basic thought of our RKNN-TSVM on a toy dataset. In this toy example, the hyperplanes of the proposed method are closer to the highly dense samples than WLTSVM. It can be observed that our RKNN-TSVM is potentially more robust to the outliers and noisy samples.

After finding the KNNs of all the samples, the weight matrix of class $+1$ (i.e. $W^{(1)}_{s_ij}$) and the margin points of class $-1$ (i.e. $f^{(2)}_{j}$) are obtained. The regularized primal problems of the proposed method are expressed as follows:
\begin{equation}\label{eq:24}
\begin{split}
\mathop{{ min}}\limits_{w_{1} ,b_{1}} \quad & \frac{1}{2}\sum\limits_{i=1}^{n_{1}}{d^{(1)}_{i}(w^{T}_{1}x^{(1)}_{i}+b_{1})^{2}}+c_{1}e_{2}^{T}\xi+\frac{c_{2}}{2}(\left\|w_{1}\right\|^{2}+b^{2}_{1}) \\
\textrm{s.t. } \quad & -f^{(2)}_{j}(w^{T}_{1}x^{(2)}_{j}+b_{1})+\xi_{j} \ge f^{(2)}_{j} \\
& \xi_{j} \ge 0,\quad j=1,...,n_{2}
\end{split}
\end{equation}
\begin{equation}\label{eq:25}
\begin{split}
\mathop{{ min}}\limits_{w_{2} ,b_{2}} \quad & \frac{1}{2}\sum\limits_{i=1}^{n_{2}}{d^{(2)}_{i}(w^{T}_{2}x^{(2)}_{i}+b_{2})^{2}}+c_{1}e_{1}^{T}\eta+\frac{c_{3}}{2}(\left\|w_{2}\right\|^{2}+b^{2}_{2}) \\
\textrm{s.t. } \quad & f^{(1)}_{j}(w^{T}_{2}x^{(1)}_{j}+b_{2})+\eta_{j} \ge f^{(1)}_{j} \\
& \eta_{j} \ge 0,\quad j=1,...,n_{1}
\end{split}
\end{equation}

\noindent where $d^{(1)}_{j}$ denotes the weight of the sample $x^{(1)}_{j}$ which is given by
\begin{equation}\label{eq:41}
d^{(1)}_{j}=\sum^{n_1}_{i=1} W^{(1)}_{s,ij},j=1,2,\dots,n_1
\end{equation}
\noindent $c_{1},c_{2},c_{3} \geq 0$ are positive parameters. $\xi $ and $\eta$ are nonnegative slack variables, both $e_{1}$ and $e_{2}$ are column vectors of ones of $n_{1}$ and $n_{2}$ dimensions, respectively.

The difference  between the primal problems of the proposed method and the existing KNN-based TSVM classifiers \cite{Ye2012, Pan2015, xu2016k} are as follows:
\begin{enumerate}
	\item Unlike WLTSVM, the value of $d^{(1)}_{j}$ depends on the distance of sample $x^{(1)}_{j}$ from its k-nearest neighbors. Therefore, the bigger the value of $d^{(1)}_{j}$, the higher dense is the sample $x^{(1)}_{j}$.
	\item Different from these classifiers, a stabilizer $\frac{c_{2}}{2}(\left\|w_{1}\right\|^{2}+b^{2}_{1})$ is added to the primal problems of (\ref{eq:24}) and (\ref{eq:25}). This makes the learning rules of our proposed method stable. In addition, the tradeoff between overfitting and generalization is dependent upon the parameters $c_{2}$ and $c_{3}$.
\end{enumerate}

Moreover, the proposed method also inherits the advantages of the existing KNN-based TSVM classifiers which are as follows:
\begin{enumerate}
	\item The optimization problems (\ref{eq:24}) and (\ref{eq:25}) are convex QPPs which have globally optimal solution.
	\item Similar to these classifiers, the computational complexity of the proposed method was reduced by only keeping the possible SVs (margin points) in the constraints.
\end{enumerate}

To solve the optimization problem (\ref{eq:24}), the Lagrangian function is given by:

\begin{align}\label{eq:26}
\begin{aligned}
L_{1}(w_{1}&,b_{1},\xi, \alpha, \gamma)= \\
&\frac{1}{2}\sum\limits_{i=1}^{n_{1}}{d^{(1)}_{j}(w^{T}_{1}x^{(1)}_{j}+b_{1})^{2}}+c_{1}e_{2}^{T}\xi+\frac{c_{2}}{2}(\left\|w_{1}\right\|^{2}+b^{2}_{1}) \\
& -\sum\limits_{j=1}^{n_{2}}{\alpha_{j}(-f^{(2)}_{j}(w^{T}_{1}x^{(2)}_{j}+b_{1})+\xi_{j} - f^{(2)}_{j})} - \gamma^{T}\xi
\end{aligned}
\end{align}

\noindent where $\alpha=(\alpha_{1}, \alpha_{2}, \dots,\alpha_{n_{2}})^{T}$ and $\gamma=(\gamma_{1}, \gamma_{2}, \dots, \gamma_{n_{2}})^{T}$ are the vectors of Lagrangian multipliers. By differentiating the Lagrangian function $L_{1}$ (\ref{eq:26}) with to respect to $w_{1}$, $b_{1}$, $\xi$, we can obtain the following the Karush-Kuhn-Tucker (KKT) conditions:
\begin{align}
\label{eq:27}
\begin{split}
\frac{\partial L_{1}}{\partial w_{1}}&= \sum\limits_{i=1}^{n_{1}}{d^{(1)}_{i}x^{(1)}_{i}(w^{T}_{1}x^{(1)}_{i} + b_{1})} + c_{2}w_{1} \\
&+ \sum\limits_{j=1}^{n_{2}}{\alpha_{j}f_{j}^{(2)}x_{j}^{(2)}}=0,
\end{split}\\
\label{eq:28}
\begin{split}
\frac{\partial L_{1}}{\partial b_{1}}&= \sum\limits_{i=1}^{n_{1}}{d^{(1)}_{i}(w^{T}_{1}x^{(1)}_{i} + b_{1})} + c_{2}b_{1} +\sum\limits_{j=1}^{n_{2}}{\alpha_{j}f_{j}^{(2)}}=0,
\end{split}\\
\label{eq:29}
\begin{split}
\frac{\partial L_{1}}{\partial \xi}&= c_{1}e_{2} - \alpha - \gamma =0,
\end{split}\\
\label{eq:30}
\begin{split} 
\alpha \geq 0,\, \gamma \geq 0.
\end{split}
\end{align}

\noindent Arranging Eqs. (\ref{eq:27}) and (\ref{eq:28}) in their matrix forms, we get the following equations:
\begin{align}
\label{eq:31}
\begin{split}
A^{T}D(Aw_{1}+e_{1}b_{1})+c_{2}w_{1}+B^{T}F\alpha=0,
\end{split}\\
\label{eq:32}
\begin{split}
e^{T}_{1}D(Aw_{1}+e_{1}b_{1}) + c_{2}b_{1}+e^{T}_{2}F\alpha=0,
\end{split}
\end{align}

\noindent where $D=diag(d^{(1)}_{1},d^{(1)}_{2},\dots,d^{(1)}_{n_{1}})$ (here, $d^{(1)}_{j} \geq 0$, $j=1,2,\dots,n_{1}$) and $F=diag(f^{(2)}_{1},f^{(2)}_{2},\dots,f^{(2)}_{n_{2}})$ are diagonal matrices. Obviously, $f^{(2)}_{j}$($j=1,2,\dots,n_{2}$) is either 0 or 1. Since $\gamma \geq 0$, from (\ref{eq:29}) we have
\begin{equation} \label{eq:33}
0e_{2} \leq \alpha \leq c_{1}e_{2}
\end{equation}

Next, combining (\ref{eq:31}) and (\ref{eq:32}) leads to the following equation
\begin{equation}\label{eq:34}
([A^{T}\ e^{T}_{1}]D[A\ e_{1}] + c_{2}I)[w_{1}\ b_{1}]^{T} + [B^{T}\ e^{T}_{2}]F\alpha = 0.
\end{equation}
\noindent where $I$ is an identity matrix of appropriate dimensions. Defining $H=[A\,e_{1}]$ and $G=[B\,e_{2}]$, the Eq.(\ref{eq:34}) can be rewritten as
\begin{align}\label{eq:35}
\begin{aligned}
&(H^{T}DH+c_{2}I)\left[ \begin{matrix}
{{w}_{1}}  \\
{{b}_{1}}  \\
\end{matrix} \right] + G^{T}F\alpha=0. \\
&i.e., \left[ \begin{matrix}
{{w}_{1}}  \\
{{b}_{1}}  \\
\end{matrix} \right] = -(H^{T}DH+c_{2}I)^{-1}G^{T}F\alpha
\end{aligned}
\end{align}

Using (\ref{eq:26}) and the above KKT conditions, the Wolfe dual of (\ref{eq:24}) is derived as follows:
\begin{equation}\label{eq:36}
\begin{split}
\mathop{{max}}\limits_{\alpha} \quad & e_{2}^{T}F\alpha-\frac{1}{2}{{\alpha }^{T}}(F^{T}G){(H^{T}DH+c_{2}I)^{-1}}{({G}^{T}F)}\alpha   \\
\textrm{s.t. } \quad & 0{{e}_{2}}\le \alpha \le {c_{1}}{{e}_{2}}
\end{split}
\end{equation}

One can notice that the parameter $c_{2}$ in the dual problem (\ref{eq:36}) can be replaced by $\varepsilon$, $\varepsilon > 0$. However, the parameter $\varepsilon$ is a very small positive scalar ($\varepsilon=1e-8$) for avoiding matrix singularity, whereas $c_2$ is a hyper-parameter which determines the tradeoff between overfitting and generalization \cite{Shao2011}. 

Similarly, the Lagrangian function of the primal problem (\ref{eq:25}) is defined as follows:
\begin{align}\label{eq:37}
\begin{aligned}
L_{2}(w_{2}&,b_{2},\eta, \beta, \nu)= \\
&\frac{1}{2}\sum\limits_{i=1}^{n_{2}}{d^{(2)}_{j}(w^{T}_{2}x^{(2)}_{j}+b_{2})^{2}}+c_{1}e_{1}^{T}\eta+\frac{c_{3}}{2}(\left\|w_{2}\right\|^{2}+b^{2}_{2}) \\
& -\sum\limits_{i=1}^{n_{1}}{\beta_{j}(f^{(1)}_{j}(w^{T}_{2}x^{(2)}_{j}+b_{2})+\eta_{j} - f^{(1)}_{j})} - \nu^{T}\eta
\end{aligned}
\end{align}

\noindent where $\beta=(\beta_{1}, \beta_{2}, \dots,\beta_{n_{1}})^{T}$ and $\nu=(\nu_{1}, \nu_{2}, \dots,\nu_{n_{1}})^{T}$ are the vectors of Lagrangian multipliers. After differentiating the Lagrangian function (\ref{eq:37}) with respect to $w_{2}$,$b_{2}$ and $\eta$, the Wolfe dual of (\ref{eq:25}) is obtained as follows:
\begin{equation}\label{eq:38}
\begin{split}
\mathop{{max}}\limits_{\beta} \quad & e_{1}^{T}P\beta-\frac{1}{2}{{\beta}^{T}}(P^{T}H){(G^{T}QG+c_{3}I)^{-1}}{({H}^{T}P)}\beta   \\
\textrm{s.t. } \quad & 0{{e}_{1}}\le \beta \le {c_{1}}{{e}_{1}}
\end{split}
\end{equation}

\noindent where $Q=diag(d^{(2)}_{1},d^{(2)}_{2},\dots,d^{(2)}_{n_{2}})$ (i.e. the weight matrix of class $-1$) and $P=diag(f^{(1)}_{1},f^{(1)}_{2},\dots,f^{(1)}_{n_{1}})$ (i.e. the weight matrix of class $+1$) are diagonal matrices. $f^{(1)}_{j}$ is either $0$ or $1$. Furthermore, it can be observed from the dual QPPs (\ref{eq:36}) and (\ref{eq:38}) that the computational complexity in the learning phase of the proposed method is affected by the number of margin points.

Once the dual QPP (\ref{eq:38}) is solved, we can obtain the following augmented vector.
\begin{equation} \label{eq:39}
\left[ \begin{matrix}
{{w}_{2}}  \\
{{b}_{2}}  \\
\end{matrix} \right] = (G^{T}QG+c_{3}I)^{-1}H^{T}P\beta
\end{equation}

Once the augmented vectors of (\ref{eq:35}) and (\ref{eq:39}) are obtained from the solutions of (\ref{eq:36}) and (\ref{eq:38}), a new testing sample $x \in \mathbb{R}^{d}$ is assigned to class $i(i=-1,+1)$ depending on which of the two hyperplanes it lies closest to. The decision function of the proposed method is given by
\begin{equation}\label{eq:40}
d(x) =
\begin{cases}
+1, & \textrm{if } \frac{\left| {{x}^{T}}{{w}_{1}}+{{b}_{1}} \right|}{\left\|w_1\right\|} < \frac{\left| {{x}^{T}}{{w}_{2}}+{{b}_{2}} \right|}{\left\|w_2\right\|} \\
-1, & \textrm{otherwise}.
\end{cases}
\end{equation}

\noindent where $\left|\ .\ \right|$ denotes the absolute value. For the sake of clearness, we explicitly state our linear RKNN-TSVM algorithm.

\begin{algo1}
	Linear RKNN-TSVM classifier
	
	Given a training set $T$ and the number of nearest neighbors $k$. The linear RKNN-TSVM can be obtained using the following steps:
	
	\begin{enumerate}
		\item To obtain the set $Nea(x_{j})$, find the $k$-nearest neighbor of each sample $x_{j} \in T$ using either FSA or LDMDBA algorithm.
		\item Define the weight matrices $W_s$ and $W_d$ for classes $+1$ and $-1$ using (\ref{eq:19}) and (\ref{eq:20}).
		\item Construct the diagonal matrices $D$, $Q$, $F$ and $P$ using (\ref{eq:41}) and (\ref{eq:23}).
		\item Construct the input matrices $A \in \mathbb{R}^{n_1 \times d}$ and $B \in \mathbb{R}^{n_2 \times d}$. Also define $H=[A\,e_{1}]$ and $G=[B\,e_{2}]$.
		\item Select parameters $c_1$, $c_2$ and $c_3$. These parameters are usually selected based on validation.
		\item Obtain the optimal solutions $\alpha$ and $\beta$ by solving the convex QPPs (\ref{eq:36}) and (\ref{eq:38}), respectively.
		\item Determine the parameters of two non-parallel hyperplanes using (\ref{eq:35}) and (\ref{eq:39}).
		\item Calculate the perpendicular distance of a new testing sample $x \in \mathbb{R}^{d}$ from the two hyperplanes. Then assign the test sample $x$ to $i(i=+1,-1)$ using (\ref{eq:40}).
	\end{enumerate}
\end{algo1}

\begin{rmk}
	In order to obtain the augmented vectors of (\ref{eq:35}) and (\ref{eq:39}), two matrix inversion $(H^{T}DH+c_{2}I)^{-1}$ and $(G^{T}QG+c_{3}I)^{-1}$ of size $(d+1)\times(d+1)$ are required, where $d$ is much smaller than the total number of samples in the training set (i.e. $d \ll n$).
\end{rmk}

\begin{rmk}
	It should be noted that the matrices $(H^{T}DH+c_{2}I)$ and $(G^{T}QG+c_{3}I)$ are positive definite matrices due to stabilizer term. Therefore, the proposed method is stable and avoids the possible ill-conditioning of $H^{T}DH$ and $G^{T}QG$.
\end{rmk}

\subsection{Nonlinear case}\label{sec:10}
In the real world, a linear kernel cannot always separate most of the classification tasks. To make nonlinear types of problems separable, the samples are mapped to a higher dimensional feature space. Thus, we extend our RKNN-TSVM to nonlinear case by considering the following kernel-generated surfaces:
\begin{equation}\label{eq:42}
K(x){{\mu}_{1}}+{{b}_{1}}=0, \quad \textrm{and} \quad K(x){{\mu}_{2}}+{{b}_{2}}=0
\end{equation}
\noindent where
\begin{equation}\label{eq:43}
K(x) = [K(x_{1},x),K(x_{2},x),\dots,K(x_{n},x)]^{T}
\end{equation}

\noindent and $K(.)$ stands for an arbitrary kernel function. The primal optimization problems of nonlinear RKNN-TSVM can be reformulated as follows:
\begin{equation}\label{eq:44}
\begin{split}
\mathop{{ min}}\limits_{\mu_{1} ,b_{1}} \quad & \frac{1}{2}\sum\limits_{i=1}^{n_{1}}{d^{(1)}_{i}(\mu^{T}_{1}K(x^{(1)}_{i})+b_{1})^{2}}+c_{1}e_{2}^{T}\xi \\
&+\frac{c_{2}}{2}(\left\|\mu_{1}\right\|^{2}+b^{2}_{1}) \\
\textrm{s.t. } \quad & -f^{(2)}_{j}(\mu^{T}_{1}K(x^{(2)}_{j}) + b_{1})+\xi_{j} \ge f^{(2)}_{j} \\
& \xi_{j} \ge 0,\quad j=1,...,n_{2}
\end{split}
\end{equation}
\begin{equation}\label{eq:45}
\begin{split}
\mathop{{ min}}\limits_{\mu_{2} ,b_{2}} \quad & \frac{1}{2}\sum\limits_{i=1}^{n_{2}}{d^{(2)}_{i}(\mu^{T}_{2}K(x^{(2)}_{i})+b_{2})^{2}}+c_{1}e_{1}^{T}\eta \\
&+\frac{c_{3}}{2}(\left\|\mu_{2}\right\|^{2}+b^{2}_{2}) \\
\textrm{s.t. } \quad & f^{(1)}_{j}(\mu^{T}_{2}K(x^{(1)}_{j})+b_{2})+\eta_{j} \ge f^{(1)}_{j} \\
& \eta_{j} \ge 0,\quad j=1,...,n_{1}
\end{split}
\end{equation}

\noindent where $c_{1},c_{2},c_{3}$ are parameters, $\xi$ and $\eta$ are the slack vectors, $d_{j}$ and $f_{j}$ are defined as in the linear case. However, the standard Euclidean metric and the distance are computed as the higher dimensional feature space instead of input space in the linear case.

Similar to the linear case, the Lagrangian function of the primal optimization problem (\ref{eq:44}) is defined as follows:

\begin{align}\label{eq:46}
\begin{aligned}
L_{1}(&\mu_{1},b_{1},\xi, \alpha, \gamma)= \\
&\frac{1}{2}\sum\limits_{i=1}^{n_{1}}{d^{(1)}_{i}(\mu^{T}_{1}K(x^{(1)}_{i})+b_{1})^{2}}+c_{1}e_{2}^{T}\xi+\frac{c_{2}}{2}(\left\|\mu_{1}\right\|^{2}+b^{2}_{1}) \\
& -\sum\limits_{j=1}^{n_{2}}{\alpha_{j}(-f^{(2)}_{j}(\mu^{T}_{1}K(x^{(2)}_{j})+b_{1})+\xi_{j} - f^{(2)}_{j})} - \gamma^{T}\xi
\end{aligned}
\end{align}

\noindent where $\alpha=(\alpha_{1}, \alpha_{2}, \dots,\alpha_{n_{2}})^{T}$ and $\gamma=(\gamma_{1}, \gamma_{2}, \dots, \gamma_{n_{2}})^{T}$ are the vectors of Lagrangian multipliers. The KKT conditions for $\mu_{1}$, $b_{1}$, $\xi$ and $\alpha$, $\gamma$ are given by

\begin{align}
\label{eq:47}
\begin{split}
\frac{\partial L_{1}}{\partial \mu_{1}}&= \sum\limits_{i=1}^{n_{1}}{d^{(1)}_{i}K(x^{(1)}_{i})(\mu^{T}_{1}K(x^{(1)}_{i}) + b_{1})} + c_{2}\mu_{1} \\
&+ \sum\limits_{j=1}^{n_{2}}{\alpha_{j}f_{j}^{(2)}K(x_{j}^{(2)})}=0,
\end{split} \\
\label{eq:48}
\begin{split}
\frac{\partial L_{1}}{\partial b_{1}}&= \sum\limits_{i=1}^{n_{1}}{d^{(1)}_{i}(\mu^{T}_{1}K(x^{(1)}_{i}) + b_{1})} + c_{2}b_{1} \\ &+\sum\limits_{j=1}^{n_{2}}{\alpha_{j}f_{j}^{(2)}}=0,
\end{split} \\
\label{eq:49}
\begin{split}
\frac{\partial L_{1}}{\partial \xi}&= c_{1}e_{2} - \alpha - \gamma =0,
\end{split} \\
\label{eq:50}
\begin{split}
\alpha \geq 0,\, \gamma \geq 0.
\end{split}
\end{align}

\noindent Arranging Eqs. (\ref{eq:47}) and (\ref{eq:48}) in their matrix forms, we obtain
\begin{align}
\label{eq:51}
\begin{split}
K(A)^{T}D(K(A)\mu_{1}+e_{1}b_{1})+c_{2}\mu_{1}+K(B)^{T}F\alpha=0,
\end{split} \\
\label{eq:52}
\begin{split}
e^{T}_{1}D(K(A)\mu_{1}+e_{1}b_{1}) + c_{2}b_{1}+e^{T}_{2}F\alpha=0,
\end{split}
\end{align}

\noindent where $K(A)$ and $K(B)$ are the kernel matrices of sizes $n_{1} \times n$ and $n_{2} \times n$, respectively ($n=n_{1}+n_{2}$). Since $\gamma \geq 0$, from (\ref{eq:50}) we have
\begin{equation} \label{eq:53}
0e_{2} \leq \alpha \leq c_{1}e_{2}
\end{equation}

Similarly, combining (\ref{eq:51}) and (\ref{eq:52}) leads to
\begin{equation}\label{eq:54}
([K(A)^{T}\ e^{T}_{1}]D[K(A)\ e_{1}] + c_{2}I)[\mu_{1}\ b_{1}]^{T} + [K(B)^{T}\ e^{T}_{2}]F\alpha = 0.
\end{equation}

Let $R=[K(A)\,e_{1}]$ and $S=[K(B)\,e_{2}]$, the Eq.(\ref{eq:54}) can be rewritten as

\begin{equation}\label{eq:55}
\left[ \begin{matrix}
{{\mu}_{1}}  \\
{{b}_{1}}  \\
\end{matrix} \right] = -(R^{T}DR+c_{2}I)^{-1}S^{T}F\alpha
\end{equation}

Then we obtain the Wolfe dual of (\ref{eq:44})
\begin{equation}\label{eq:56}
\begin{split}
\mathop{{max}}\limits_{\alpha} \quad & e_{2}^{T}F\alpha-\frac{1}{2}{{\alpha }^{T}}(F^{T}S){(R^{T}DR+c_{2}I)^{-1}}{({S}^{T}F)}\alpha   \\
\textrm{s.t. } \quad & 0{{e}_{2}}\le \alpha \le {c_{1}}{{e}_{2}}
\end{split}
\end{equation}

In a similar manner, we can obtain the Wolfe dual of the primal optimization problem (\ref{eq:45}) by reversing the roles of $K(A)$ and $K(B)$ in (\ref{eq:56}):
\begin{equation}\label{eq:57}
\begin{split}
\mathop{{max}}\limits_{\beta} \quad & e_{1}^{T}P\beta-\frac{1}{2}{{\beta}^{T}}(P^{T}R){(S^{T}QS+c_{3}I)^{-1}}{({R}^{T}P)}\beta   \\
\textrm{s.t. } \quad & 0{{e}_{1}}\le \beta \le {c_{1}}{{e}_{1}}
\end{split}
\end{equation}

Once the dual QPP (\ref{eq:57}) is solved, we will obtain
\begin{equation}\label{eq:58}
\left[ \begin{matrix}
{{\mu}_{2}}  \\
{{b}_{2}}  \\
\end{matrix} \right] = (S^{T}QS+c_{3}I)^{-1}R^{T}P\beta
\end{equation}

\noindent Here, the specifications of the matrices $D$, $F$, $P$ and $Q$ are analogous to the linear case. In the nonlinear case, a new testing sample $x$ is assigned to class $i(i=-1,+1)$ depending on which of the two hypersurfaces it lies closest to. The decision function of nonlinear RKNN-TSVM is as follows
\begin{equation}\label{eq:59}
d(x) =
\begin{cases}
+1, & \textrm{if } \frac{\left|{K({x})}{{\mu}_{1}}+{{b}_{1}} \right|}{\left\|\mu_1\right\|} < \frac{\left| {K({x})}{{\mu}_{2}}+{{b}_{2}} \right|}{\left\|\mu_2\right\|} \\
-1, & \textrm{otherwise}.
\end{cases}
\end{equation}

We now state explicitly our nonlinear RKNN-TSVM algorithm.
\begin{algo1}
	Nonlinear RKNN-TSVM classifier
	
	Given a training set $T$ and the number of nearest neighbors $k$. The nonlinear RKNN-TSVM can be obtained using the following steps:
	\begin{enumerate}
		\item Choose a kernel function $K$.
		\item In the high dimensional feature space, find the $k$-nearest neighbor of each sample $x_{j} \in T$ using either FSA or LDMDBA algorithm.
		\item Define the weight matrices $W_s$ and $W_d$ for classes $+1$ and $-1$ using (\ref{eq:19}) and (\ref{eq:20}).
		\item Construct the diagonal matrices $D$, $Q$, $F$ and $P$ using (\ref{eq:41}) and (\ref{eq:23}).
		\item Construct the input matrices $A \in \mathbb{R}^{n_1 \times d}$ and $B \in \mathbb{R}^{n_2 \times d}$. Also define $R=[K(A)\,e_{1}]$ and $S=[K(B)\,e_{2}]$.
		\item Select parameters $c_1$, $c_2$ and $c_3$. These parameters are usually selected based on validation.
		\item Obtain the optimal solutions $\alpha$ and $\beta$ by solving the convex QPPs (\ref{eq:56}) and (\ref{eq:57}), respectively.
		\item Determine the parameters of two hypersurfaces using (\ref{eq:55}) and (\ref{eq:58}).
		\item Calculate the perpendicular distance of a new testing sample $x \in \mathbb{R}^{d}$ from the two hypersurfaces. Then assign the test sample $x$ to $i(i=+1,-1)$ using (\ref{eq:59}).
	\end{enumerate}
\end{algo1}

\begin{rmk}
	It can be noted that our nonlinear RKNN-TSVM requires the inversion of a matrix of size $(n \times 1) \times (n \times 1)$ twice. In order to reduce the computational cost, two approaches can be applied to our nonlinear RKNN-TSVM:
	\begin{enumerate}
		\item The rectangular kernel technique \cite{Mangasarian} can be used to reduce the dimensionality.
		\item The Sherman-Morisson-Woodbury (SMW) formula \cite{Golub2012} can be utilized to compute the matrix inverses of smaller dimension than $(n \times 1) \times (n \times 1)$.
	\end{enumerate}
\end{rmk}

\section{Analysis of algorithm and a fast iterative algorithm clipDCD}\label{sec:11}
\subsection{The framework of RKNN-TSVM}\label{sec:29}
Similar to other KNN-based TSVM classifiers, the output model of the proposed method is created by performing 3 steps. However, each step in the framework of our RKNN-TSVM was improved. These steps are explained as follows:
\begin{enumerate}
	\item In the first step, the KNNs of all the training samples are computed. Nonetheless, the LDMDBA algorithm was employed to accelerate the process of KNN finding.
	\item After KNN computation, the intra-class matrix $W_{s,ij}$ and inter-class matrix $W_{d,ij}$ are obtained. Using $W_{s}$ matrix, weight is given to each sample with respect to the distance from its nearest neighbors. Finally, margin points are determined using inter-class $W_{d}$ matrix.
	\item In order to obtain the output model, two dual optimization problems and two systems of linear equations are solved. The third step was improved by considering the SRM principle in the optimization problems of the proposed method.
\end{enumerate}

\begin{figure*}[!t]
	\centering
	\begin{tikzpicture}[node distance=2cm]
	\node (input) [io] {Training samples};
	\node (knn)  [process, below of =input] {Find KNNs of all the training samples using LDMDBA algorithm.};
	\node (k) [para, below of=knn] {$k$};
	\node (weight)  [process, right of = knn, xshift=2cm] {Compute the weights of samples and extract margin points.};
	\node (cl)  [process, right of = weight, xshift=2cm] {Train the RKNN-TSVM classifier.};
	\node (param) [para, above of=cl] {$c_1, c_2, c_3$};
	\node (model)  [process, right of =cl, xshift=2cm] {The output model. \\ ${{x}^T}{{w}_{1}}+{{b}_{1}}=0$ \\ ${{x}^T}{{w}_{2}}+{{b}_{2}}=0$};
	
	\draw [arrow] (input) -- (knn);
	\draw [arrow] (knn) -- (weight);
	\draw [arrow] (weight) -- (cl);
	\draw [arrow] (cl) -- (model);
	\draw [arrow] (param) --(cl);
	\draw [arrow] (k) -- (knn);
	
	\end{tikzpicture}
	\caption{Overview of steps performed by the proposed method.}
	\label{fig:8}
\end{figure*}
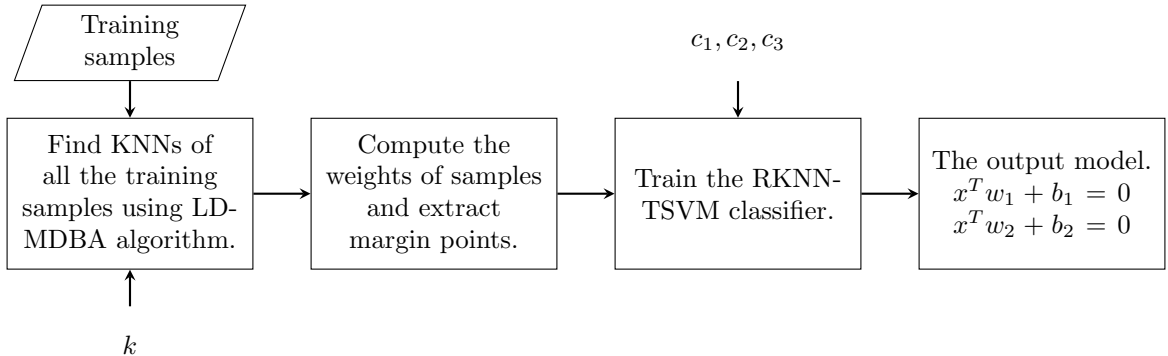

Fig. \ref{fig:8} shows the overview of steps performed by the proposed method.

\subsection{The computational complexity of RKNN-TSVM}\label{sec:28}
The major computation in our RKNN-TSVM involves two steps:
\begin{enumerate}
	\item To obtain the output model, the proposed method needs to solve two smaller sized dual QPPs. However, the size of dual problems is affected by the number of extracted margin points. After all, optimization of RKNN-TSVM costs around $\mathcal{O}(n^{3})$.
	\item To compute the weight matrices, RKNN-TSVM has to find $k$-nearest neighbors for all $n$ training samples. By using the FSA algorithm, the KNN step costs about $\mathcal{O}(n^{2}\log{n})$. To reduce the computational cost of the KNN step, the proposed method employs a fast KNN method, LDMDBA algorithm which has a computational complexity of $\mathcal{O}(\log{d}n\log{n})$.
\end{enumerate}
Thus, the overall computational complexity of RKNN-TSVM is about $\mathcal{O}(n^{3}+\log{d}n\log{n})$.

\subsection{Comparison with other related algorithms}\label{sec:18}
In this subsection, we compare our RKNN-TSVM with other related algorithms.

\subsubsection{Comparison with TSVM}\label{sec:19}
Compared with TSVM \cite{Khemchandani2007}, our RKNN-TSVM gives weight to the samples of each class by using the KNN graph. As a result, the hyperplanes of the proposed method are closer to the samples with greater weight and far from the margin points of the other class instead of all the samples. Moreover, the proposed method improves the computational cost of solving QPPs by only keeping the margin points in the constraints.

TSVM only considers the empirical risk which may lead to overfitting problem. However, our RKNN-TSVM implements the SRM principle by adding a stabilizer term to the objective function. As a result, the proposed method achieves better classification accuracy and generalization.

\subsubsection{Comparison with WLTSVM}\label{sec:20}
Unlike WLTSVM \cite{Ye2012}, our RKNN-TSVM gives weight to each sample with respect to the distance from its nearest neighbors. As a result, the neighbors with smaller distance were weighted more heavily than the one with greater distance. Moreover, the proposed method finds KNNs of all the samples by utilizing fast KNN method such as LDMDBA \cite{xia2015location} algorithm. This makes the learning speed of our RKNN-TSVM faster than that of WL-TSVM.

Different from WLTSVM, the optimization problems of the proposed method are regularized and stable. Hence two parameters $c_2$ and $c_3$ can be adjusted to determine the tradeoff between overfitting and generalization. This makes our RKNN-TSVM better in terms of classification accuracy.

\subsubsection{Comparison with KNN-STSVM}\label{sec:21}
Similar to WLTSVM, KNN-STSVM \cite{Pan2015} gives weight to each sample by only counting the number of its nearest neighbors. Also, it does not consider the SRM principle which makes the classifier stable. Moreover, KNN-STSVM extracts the data distribution information in objective function by using Ward's linkage clustering method. 

In summary, KNN-STSVM consists of three steps: (1) Getting proper clusters. (2) KNN finding. (3) Solving two smaller-sized QPPs. The overall computational complexity of this classifier is around $\mathcal{O}(1/4n^{3} + d(n^{2}_{1} + n^{2}_{2}) + n^{2}\log{n})$. Therefore, it cannot handle large scale datasets. 

\subsection{The limitation of our RKNN-TSVM}\label{sec:22}
We should acknowledge that our RKNN-TSVM has the following limitations:
\begin{enumerate}
	\item Because of solving two systems of linear equations, the matrix inverse operation is unavoidable in our RKNN-TSVM. The computational complexity of the matrix inverse is $\mathcal{O}(n^{3})$. This implies that the computational cost rapidly increases with the increase of matrix order.
	\item For large scale datasets, the memory consumption of the proposed method is very high. Because two nearest neighbor graphs need to be stored.
	\item Even though SRM principle boosts the classification accuracy of our RKNN-TSVM, it comes at the cost of tuning two additional parameters. In total, there are four parameters $c_{1}$, $c_{2}$, $c_{3}$, $k$ in our RKNN-TSVM which need to be adjusted. Therefore, the parameter selection of the proposed method is computationally expensive. In the experiments, we set $c_{2} = c_{3}$ to reduce the computational cost of parameter selection.
\end{enumerate}

\subsection{The scalability of RKNN-TSVM}\label{sec:27}
Similar to WLTSVM, our RKNN-TSVM introduces the selection vector $f_{j}$ to the constraint of the optimization problems. As a result, it considers only the margin points instead of all the samples for obtaining the output model. This further reduces the time complexity of the classifier. However, the proposed method has better scalability in comparison with WLTSVM. It utilizes LDMDBA algorithm to find KNNs of all the samples. This algorithm decreases the computational cost of KNN finding and makes our RKNN-TSVM more suitable for large scale datasets.

\subsection{The clipDCD algorithm}\label{sec:30}
In our RKNN-TSVM, there are four strictly convex dual QPPs to be solved: (\ref{eq:36}),(\ref{eq:38}),(\ref{eq:56}) and (\ref{eq:57}). These optimization problems can be rewritten in the following unified form:
\begin{equation}\label{eq:62}
\begin{split}
\mathop{{ min}}\limits_{\alpha} \qquad & f(\alpha)=\frac{1}{2} \alpha^{T}Q\alpha - e^{T}\alpha, \\
\textrm{s.t. } \qquad & 0 \leq \alpha \leq c.
\end{split}
\end{equation}

\noindent where $Q \in \mathbb{R}^{n \times n}$ is positive definite. For example, the matrix $Q$ in (\ref{eq:62}) can be substituted by
\begin{align*}
(F^{T}S){(R^{T}DR+ c_{2}I)^{-1}}{({S}^{T}F)}.
\end{align*}

To solve the dual QPP problem (\ref{eq:62}), a solver algorithm is required. So far, many fast training algorithms were proposed which may include but not limited to, interior-point methods \cite{sra2012optimization}, successive overrelaxation (SOR) technique \cite{mangasarian1999successive} and dual coordinate descent (DCD) algorithm \cite{Hsieh}. On the basis of DCD, Peng et al. proposed the clipping dual coordinate descent (clip\-DCD) algorithm \cite{Peng2014}.

In this paper, we employed the clipDCD algorithm \cite{Peng2014} to speed up the learning process of our RKNN-TSVM. The main characteristics of this algorithm are fast learning speed and easy implementation. The clip\-DCD is a kind of the gradient descent method. Its main idea is to orderly select and update a variable which is based on the maximal possibility-decrease strategy \cite{Peng2014}.

Unlike the DCD algorithm, this method does not consider any outer and inner iteration. That is, only one component of $\alpha$ is updated at each iteration, denoted $\alpha_{L} \rightarrow \alpha_{L} + \lambda$, $L \in \{1,\dots,n\}$ is the index. Then the objective function is defined as follows:
\begin{equation}\label{eq:63}
f(\lambda)=f(0) + \frac{1}{2}\lambda^{2}Q_{LL} - \lambda(e_{L} - \alpha^{T}Q_{.,L}).
\end{equation}

\noindent where $Q_{.,L}$ is the $L$th column of the $Q$ matrix. Setting the derivation of $\lambda$
\begin{equation}\label{eq:64}
\frac{df(\lambda)}{d\lambda}=0 \Rightarrow \lambda = \frac{(e_{L} - \alpha^{T}Q_{.,L})}{Q_{LL}}
\end{equation} 

The largest decrease on the objective value can be derived by choosing the $L$ index as:
\begin{equation}\label{eq:65}
L = \mathop{arg\,max}\limits_{i \in S}\Big\{\frac{(e_{i} - \alpha^{T}Q_{.,i})^{2}}{Q_{ii}}{} \Big\},
\end{equation}

\noindent where the index set $S$ is
\begin{equation}\label{eq:66}
\begin{split}
\small
S =\Big\{&i:\alpha_{i} > 0 \textrm{ if } \frac{e_{i} - \alpha^{T}Q_{.,i}}{Q_{ii}} < 0 \\ &\textrm{ or }\alpha_{i} < c \textrm{ if } \frac{e_{i} - \alpha^{T}Q_{.,i}}{Q_{ii}} > 0 \Big\}.
\end{split}
\end{equation}

The stopping criteria of the clipDCD algorithm is defined as follows:
\begin{equation}\label{eq:67}
\frac{(e_{L} - \alpha^{T}Q_{.,L})^{2}}{Q_{LL}} < \epsilon, \quad \epsilon > 0
\end{equation}

\noindent where the tolerance parameter $\epsilon$ is a positive small number. We set $\epsilon=10^{-5}$ in our experiments. The whole process of solving convex dual QPPs using clipDCD solver is summarized in the Algorithm \ref{algo_clip}. More information on the convergence of this algorithm and other theoretical proofs can be found in \cite{Peng2014}.

\begin{pcode}[t]
	\SetKwInOut{In}{Input}
	\SetKwInOut{Out}{Output}
	\SetKw{Kwor}{or}
	\SetKw{Kwnot}{not}
	\SetKw{KwBreak}{break}
	
	\In{$Q \in \mathbb{R}^{n \times n}$, $c$}
	\Out{The best set of Lagrange multipliers $\alpha$}
	\BlankLine
	Initialize $\alpha \leftarrow 0$\;
	$t \leftarrow 0$ \tcc*[r]{Iteration counter}
	$e \leftarrow [1,1 \dots, 1]_{n\times1}^{T}$\;
	Index set $S=\{1,2,\dots,n\}$\;
	\While{$\alpha$ is not optimal}{
		
		\For{$i \in S$}{
			
			\If{\Kwnot$(\alpha_{i} < c$ \Kwor $\frac{e_{i} - \alpha^{T}Q_{.,i}}{Q_{ii}} > 0 )$}{
				
				$S \leftarrow S - \{i\}$\;
			}
		}
		
		\tcp{Choose $L$ index}
		$L = \mathop{arg\,max}\limits_{i \in S}\Big\{\frac{(e_{i} - \alpha^{T}Q_{.,i})^{2}}{Q_{ii}}{} \Big\}$\;
		
		\tcp{Compute $\lambda$}
		$\lambda = \frac{(e_{L} - \alpha^{T}Q_{.,L})}{Q_{LL}}$\;
		
		\tcp{Update alpha value}
		$\alpha^{new}_{L} \leftarrow \alpha_{L} + \mathop{max}\{0, \mathop{min}\{\lambda,c\}\}$\;
		
		\tcp{Check the stopping criteria}
		\If{$\frac{(e_{L} - \alpha^{T}Q_{.,L})^{2}}{Q_{LL}} < \epsilon$}{
			\KwBreak\;
		}
		
	}
	
	\caption{The clipDCD solver}\label{algo_clip}
\end{pcode}

\section{Numerical experiments}\label{sec:12}
In this section, we conduct extensive experiments on several synthetic and benchmark datasets to investigate the classification accuracy and the computational cost of our RKNN-TSVM. In each subsection, the experimental results and the corresponding analysis are given.

\subsection{Experimental setup and implementation details}\label{sec:17}
For experiments with TSVM, we used LightTwinSVM\footnote{https://github.com/mir-am/LightTwinSVM} program \cite{ltsvm2019} which is an open source and fast implementation of standard TSVM classifier. All other classifiers were implemented in Python\footnote{https://www.python.org} 3.5 programming language. NumPy \cite{walt2011numpy} package was used for linear algebra operations such as matrix multiplication and inverse. Moreover, SciPy \cite{jones2014scipy} package was used for distance calculation and statistical functions. For model selection and cross-validation, Scikit-learn \cite{2011scikit} package was employed. To solve dual QPPs, the C++ implementation of clipDCD optimizer within LightTwinSVM's code was used. The LDMDBA algorithm was implemented in C++ with GNU Compiler Collection\footnote{https://gcc.gnu.org} 5.4 (GCC). Pybind 11\footnote{https://pybind11.readthedocs.io/en/stable/intro.html} was employed to create Python binding of C++ code. All the experiments were carried out on Ubuntu 16.04 LTS with an Intel Core i7 6700K CPU (4.2 GHz) and 32.0 GB of RAM.

\subsection{Parameters selection}\label{sec:13}
The classification performance of TSVM-based classifiers depends heavily on the choice of parameters. In our experiments, the grid search method is employed to find the optimal parameters. In the case of non-linear kernel, the Gaussian kernel function $K(x_{i},x_{j})=exp(-\frac{\left\|x_{i}-x_{j}\right\|}{2\sigma^{2}})$ is used as it is often employed and yields great generalization performance. The optimal value of the Gaussian kernel parameter $\sigma$ was selected over the range $\{2^{i} \mid i=-10,-9,\dots,2 \}$. The optimal value of the parameters $c_{1}$,$c_{2}$,$c_{3}$ was selected from the set $\{2^{i} \mid i=-8,-7,\dots,2 \}$. To reduce the computational cost of the parameter selection, we set $c_{1}=c_{2},c_{3}=c_{4}$ in TBSVM and $c_{2} = c_{3}$ for RKNN-TSVM. In addition, the optimal value for $k$ in RKNN-TSVM and WLTSVM was chosen from the set $\{2,3,\dots,15\}$.

\subsection{Experimental results and discussion}\label{sec:14}
In this subsection, we analyze the results of the proposed method on several synthetic and benchmark datasets from the perspective of the prediction accuracy and computational efficiency.

\begin{figure*}[!t]
	\centering
	
	\subfloat[Linear WLTSVM ($c=2^{3}, k=2$)]{%
		\includegraphics[width=0.5\columnwidth]{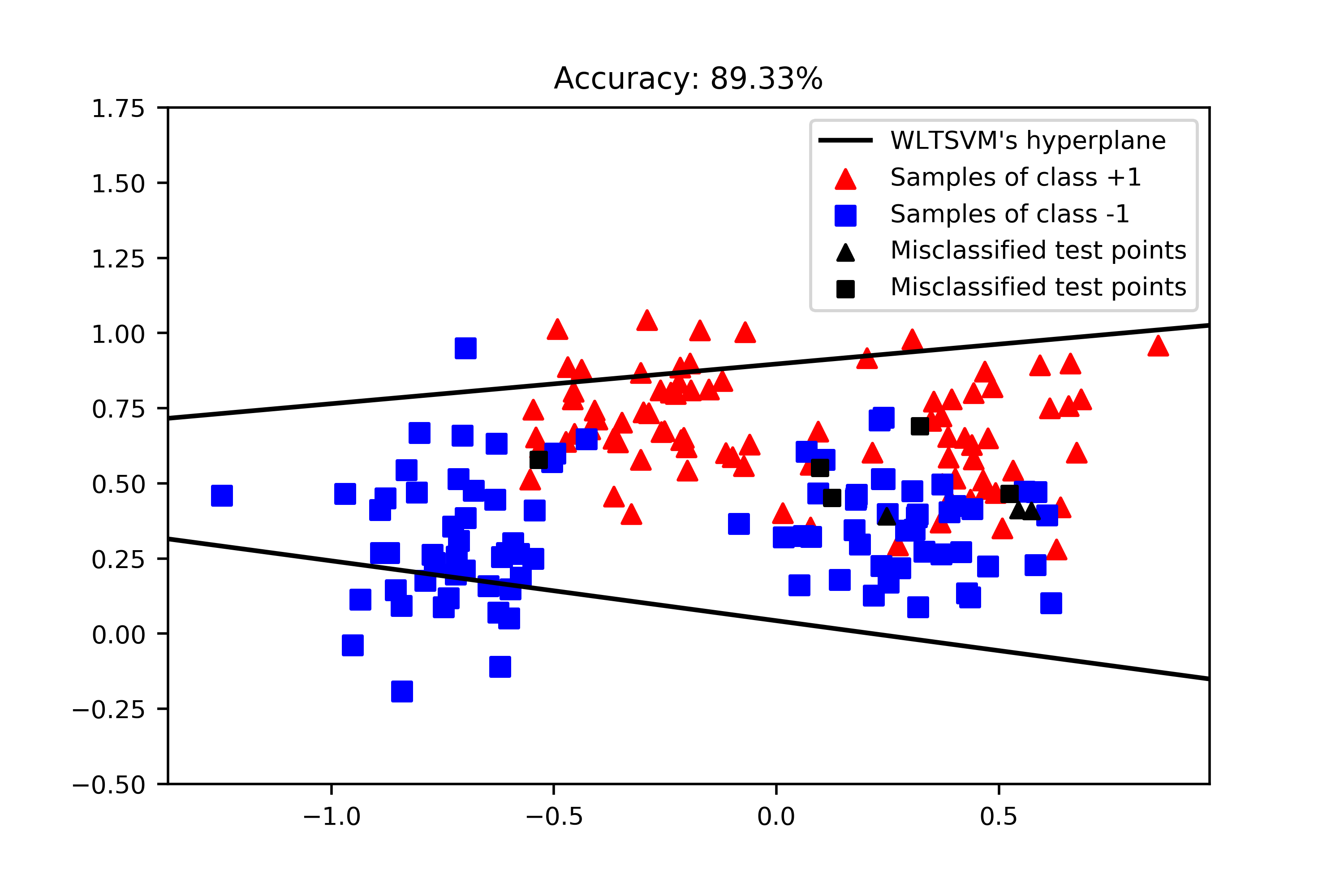}}
	\subfloat[Linear RKNN-TSVM ($c_{1}=2^{2}, c_{2}=2^{-7}, c_{3}=2^{-2}, k=6$)]{%
		\includegraphics[width=0.5\columnwidth]{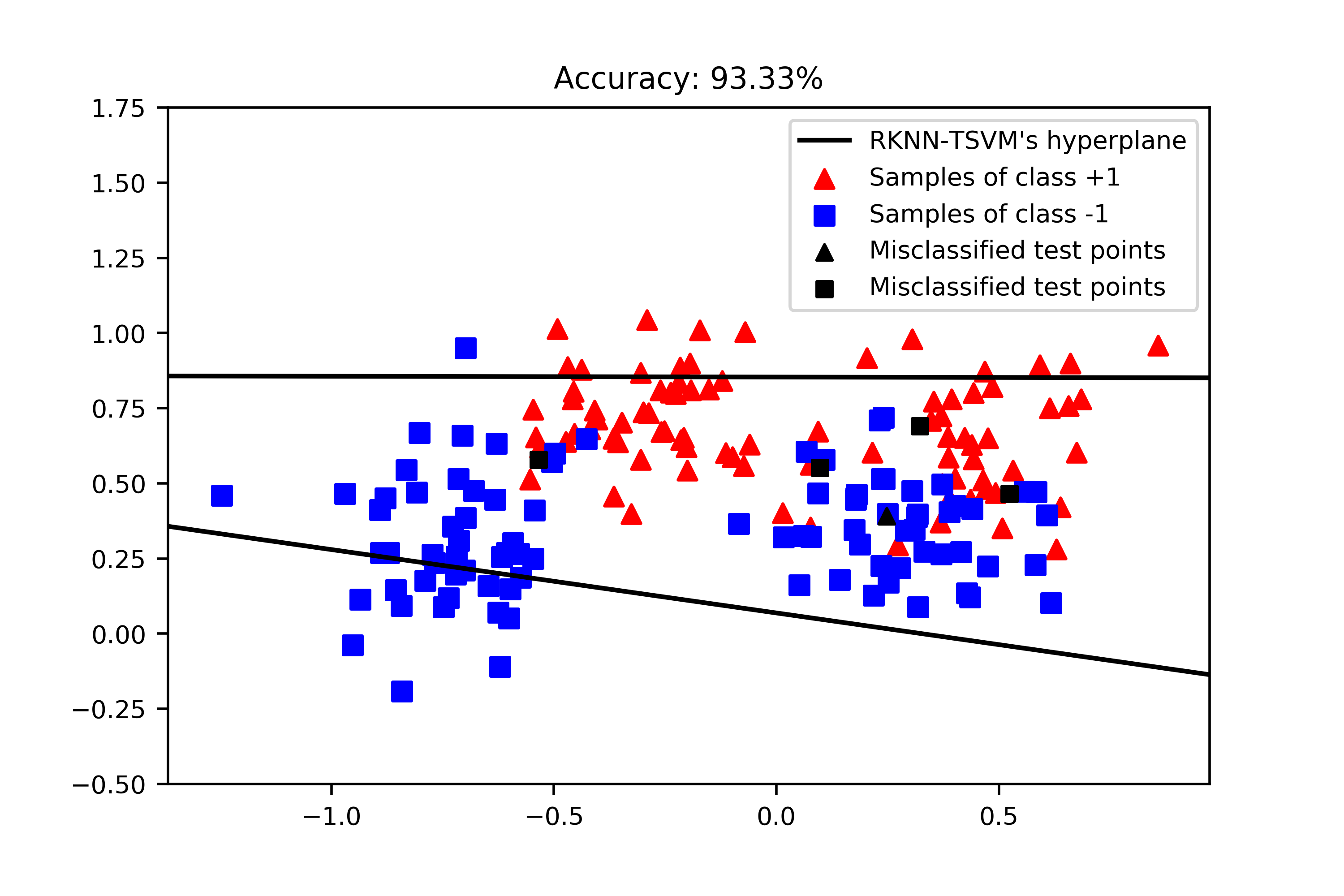}}
	
	\caption{The performance and graphical representation of WLTSVM and RKNN-TSVM on Ripley's dataset with linear kernel}\label{fig:2}
\end{figure*}
\begin{figure*}[!t]
	\centering
	
	\subfloat[Nonlinear WLTSVM ($c=2^{-3}, k=8, \sigma=2^{0}$)]{\includegraphics[width=0.5\columnwidth]{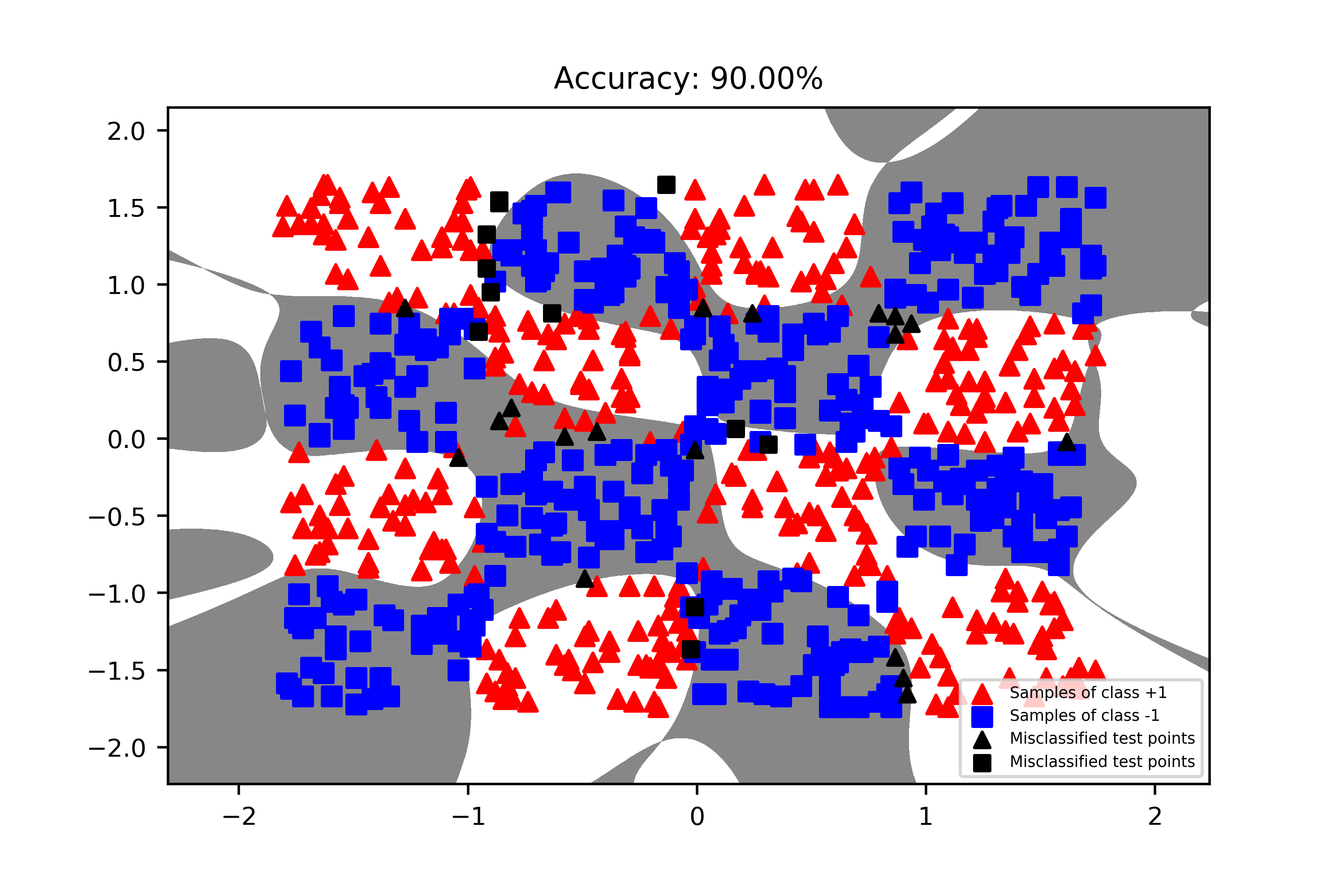}}
	\subfloat[Nonlinear RKNN-TSVM ($c_{1}=2^{-7}, c_{2}=2^{-6}, c_{3}=2^{-6}, k=10, \sigma=2^{1}$)]{\includegraphics[width=0.5\columnwidth]{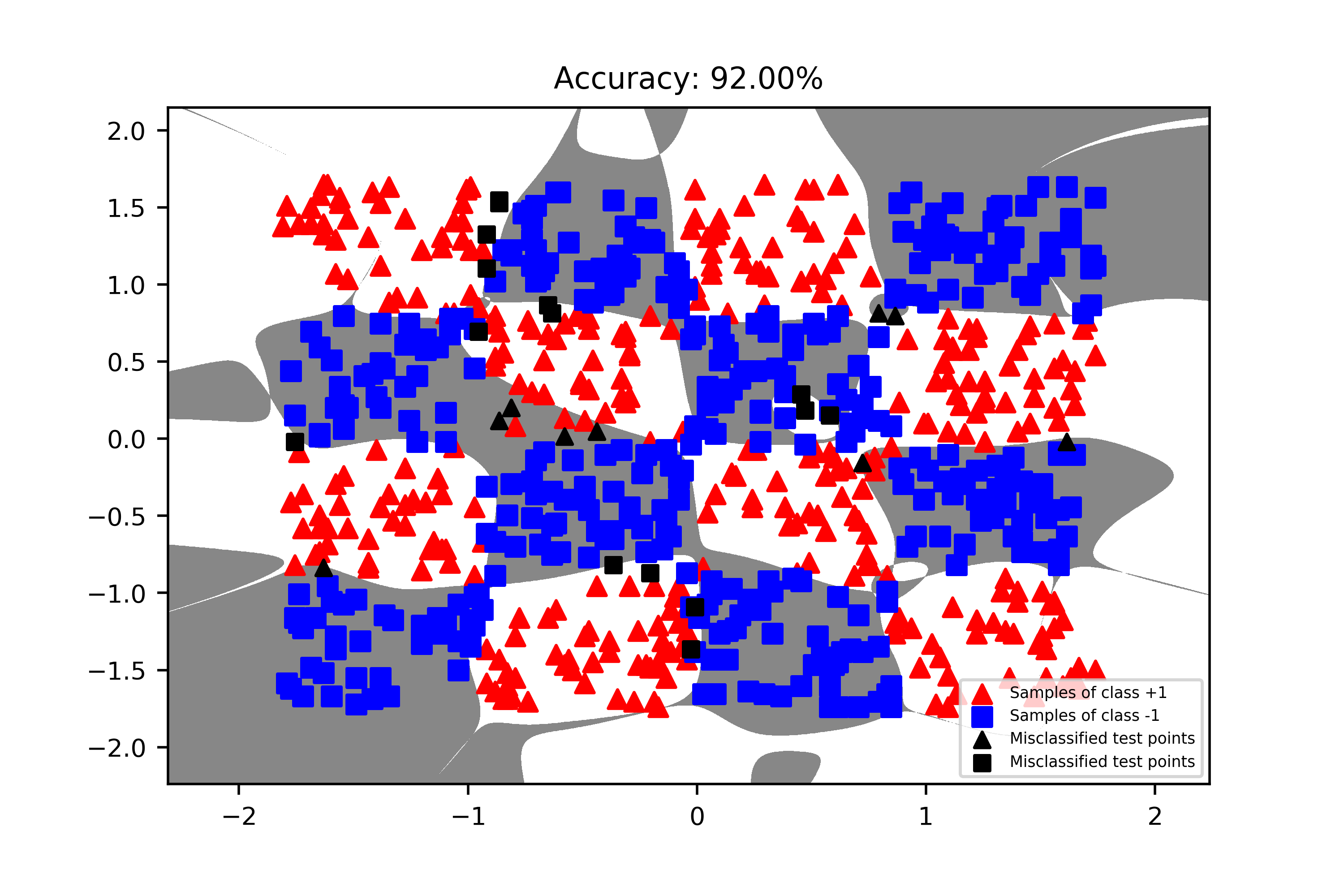}}
	
	\caption{The performance and graphical representation of WLTSVM and RKNN-TSVM on checkerboard dataset with Gaussian kernel}\label{fig:3}
\end{figure*}

\subsubsection{Synthetic datasets}\label{sec:15}
To demonstrate graphically the effectiveness of our RKNN-TSVM over WLTSVM, we conducted experiments on two artificially-generated synthetic datasets. For experiments with these datasets, 70\% of samples are randomly chosen as the training samples. 

In the first example, we consider the two dimensional Ripley's synthetic dataset \cite{ripley2007pattern} which includes 250 samples. Fig. \ref{fig:2} shows the performance and graphical representation of WLTSVM and RKNN-TSVM on Ripley's dataset with a linear kernel. By inspecting Fig. \ref{fig:2}, one can observe that our linear RKNN-TSVM obtains better classification performance and its hyperplanes are proximal to the highly dense samples. This is because the proposed method gives weight to each sample with respect to the distance from its nearest neighbors.

The second example is a two-dimensional checkerboard dataset \cite{ho1996checkerboard} which includes 1000 samples. Fig \ref{fig:3} visually displays the performance of WLTSVM and RKNN-TSVM on checkerboard dataset with Gaussian kernel. As shown in Fig. \ref{fig:3}, the accuracy of our non-linear is better than that of nonlinear WLTSVM. Because our RKNN-TSVM considers the SRM principle which improves the generalization ability. Moreover, as mentioned earlier, the proposed method gives weight based on the distance between a sample and its nearest neighbors. 

\subsubsection{Benchmark datasets}\label{sec:16}
To further validate the efficiency of the proposed method, we compare the performance of our RKNN-TSVM with TSVM, TBSVM and WLTSVM on benchmark datasets from the UCI machine learning repository\footnote{http://archive.ics.uci.edu/ml/datasets.html}. It should be noted that all the datasets were normalized such that the feature values locate in the range $\left[0,1\right]$. The characteristics of these datasets are shown in Table \ref{tab:2}.

Experiments were performed using $5$-fold cross-validation in order to evaluate the performance of these algorithms and tune parameters. More specifically, the dataset is split randomly into 5 subsets, and one of those sets is reserved as a test set. This procedure is repeated 5 times, and the average of 5 testing results is used as the performance measure.

The classification accuracy and running time of TSVM, TBSVM, WLTSVM, and RKNN-TSVM are summarized in Table \ref{tab:3}. Here, ``Accuracy'' denotes the mean value of the testing results (in \%) and the corresponding standard deviation. ``Time'' denotes the mean value of training time.

\begin{table}[!t]
	\small
	\centering
	\caption{The characteristics of benchmark datasets}
	\tabcolsep=0.11cm
	\begin{tabular}{l c c c c}
		\hline
		Datasets & \#Samples & \#Positive & \#Negative & \#Features \\
		\hline
		{Australian} & {690} & {307} & {383} & {14} \\
		{Heart-Statlog} & {270} & {120} & {150} & {13} \\
		{Bupa-Liver} & {345} & {145} & {200} & {6} \\
		{WPBC} & {198} & {47} & {151} & {33} \\
		{WDBC} & {569} & {212} & {357} & {30} \\
		{Hepatitis} & {155} & {32} & {123} & {19} \\
		{Ionosphere} & {351} & {225} & {126} & {34} \\
		{Haberman} & {306} & {225} & {81} & {3} \\
		{Pima-Indian} & {768} & {268} & {500} & {8} \\
		
		
		{Fertility} & {100} & {88} & {12} & {9} \\
		{Votes} & {435} & {267} & {168} & {16} \\
		
		\hline
	\end{tabular}
	
	\label{tab:2}
\end{table}

\begin{sidewaystable*}
	
	\small
	\centering
	\caption{Performance comparison of TSVM, TBSVM, WLTSVM and, RKNN-TSVM on benchmark datasets with Gaussian kernel. Bold value denotes the best result.}
	\ra{1.3} 
	\tabcolsep=0.13cm 
	\begin{tabular}{p{2.70cm} \acol \tcol c \acol \tcol c \acol \tcol c \acol \tcol c \acol \tcol}
		\toprule
		
		Datasets & \multicolumn{2}{c}{TSVM} && \multicolumn{2}{c}{TBSVM} && \multicolumn{2}{c}{WLTSVM} && \multicolumn{2}{c}{RKNN-TSVM(FSA)} && \multicolumn{2}{c}{RKNN-TSVM(LDMDBA)} \\
		\cmidrule{2-3} \cmidrule{5-6} \cmidrule{8-9} \cmidrule{11-12} \cmidrule{14-15}
		($n\times d$) & Accuracy(\%) & Time (s) && Accuracy(\%) & Time (s) && Accuracy(\%) & Time (s) && Accuracy(\%) & Time (s) && Accuracy(\%) & Time (s) \\
		& $(c_{1}, c_{2}, \sigma)$  &  && $(c_{1}, c_{3}, \sigma)$  & && $(c, \sigma, k)$ &  && $(c_{1}, c_{2}, \sigma, k)$  &  && $(c_{1}, c_{2}, \sigma, k)$  &  \\
		\midrule
		Australian & 87.10$\pm$3.09 & 0.066 && 87.39$\pm$3.39 & 0.062 && 86.52$\pm$3.53 & 0.144 && 87.54$\pm$3.65 & 0.147 && \textbf{87.97$\pm$3.85} & 0.226 \\
		($690\times 14$) & ($2^{-4}, 2^{-5}, 2^{-7}$) &  && ($2^{-5}, 2^{2}, 2^{-6}$) &  && ($2^{2}, 2^{-8}, 14$) &  && ($2^{1}, 2^{-3}, 2^{-9}, 2$) &  && ($2^{-4}, 2^{-3}, 2^{-6}, 5$) & \\ 
		Heart-Statlog & 84.81$\pm$2.72 & 0.010 && \textbf{85.93$\pm$2.51} & 0.013 && 83.70$\pm$1.39 & 0.023 && \textbf{85.93$\pm$3.01} & 0.023 && 85.56$\pm$2.16 & 0.028 \\
		($270\times 13$) & ($2^{0}, 2^{-1}, 2^{-10}$) &  && ($2^{0}, 2^{-3}, 2^{-10}$) &  && ($2^{0}, 2^{-7}, 12$) &  && ($2^{2}, 2^{-1}, 2^{-10}, 4$) &  && ($2^{1}, 2^{-5}, 2^{-10}, 5$) & \\ 
		Bupa-Liver & \textbf{74.78$\pm$2.35} & 0.016 && 73.62$\pm$2.13 & 0.029 && 73.91$\pm$2.05 & 0.049 && 73.91$\pm$4.30 & 0.036 && 73.91$\pm$4.58 & 0.066 \\
		($345\times 6$) & ($2^{1}, 2^{1}, 2^{-7}$) &  && ($2^{-2}, 2^{-7}, 2^{-5}$) &  && ($2^{0}, 2^{-6}, 10$) &  && ($2^{2}, 2^{-2}, 2^{-5}, 10$) &  && ($2^{2}, 2^{-3}, 2^{-5}, 7$) & \\ 
		WPBC & 79.27$\pm$5.48 & 0.017 && 78.81$\pm$7.47 & 0.012 && 78.82$\pm$8.05 & 0.016 && 80.29$\pm$3.78 & 0.013 && \textbf{80.32$\pm$3.98} & 0.028 \\
		($198\times 33$) & ($2^{-2}, 2^{-5}, 2^{-6}$) &  && ($2^{0}, 2^{-5}, 2^{-9}$) &  && ($2^{-3}, 2^{-7}, 7$) &  && ($2^{-1}, 2^{-2}, 2^{-5}, 11$) &  && ($2^{-2}, 2^{-5}, 2^{-6}, 10$) & \\ 
		WDBC & 98.24$\pm$1.36 & 0.072 && 98.24$\pm$0.78 & 0.055 && 97.54$\pm$1.02 & 0.090 && \textbf{98.59$\pm$0.70} & 0.123 && \textbf{98.59$\pm$0.70} & 0.157 \\
		($569\times 30$) & ($2^{-4}, 2^{-2}, 2^{-9}$) &  && ($2^{-5}, 2^{-7}, 2^{-8}$) &  && ($2^{1}, 2^{-7}, 8$) &  && ($2^{-3}, 2^{-4}, 2^{-6}, 6$) &  && ($2^{0}, 2^{-3}, 2^{-7}, 8$) & \\ 
		Hepatitis & 85.81$\pm$7.80 & 0.004 && 87.10$\pm$5.77 & 0.004 && 85.16$\pm$5.98 & 0.012 && 87.74$\pm$7.18 & 0.017 && \textbf{88.39$\pm$6.95} & 0.015 \\
		($155\times 19$) & ($2^{-4}, 2^{-5}, 2^{-9}$) &  && ($2^{-5}, 2^{0}, 2^{-5}$) &  && ($2^{-5}, 2^{-7}, 11$) &  && ($2^{-4}, 2^{-3}, 2^{-6}, 7$) &  && ($2^{-4}, 2^{-3}, 2^{-6}, 3$) & \\ 
		Ionosphere & 90.89$\pm$4.07 & 0.031 && 92.02$\pm$4.91 & 0.015 && 92.60$\pm$3.97 & 0.057 && \textbf{93.73$\pm$3.45} & 0.047 && 93.17$\pm$3.87 & 0.066 \\
		($351\times 34$) & ($2^{-2}, 2^{-4}, 2^{-5}$) &  && ($2^{-8}, 2^{-5}, 2^{0}$) &  && ($2^{-5}, 2^{1}, 10$) &  && ($2^{-3}, 2^{1}, 2^{0}, 5$) &  && ($2^{-5}, 2^{2}, 2^{0}, 12$) & \\ 
		Haberman & 75.46$\pm$5.06 & 0.015 && 75.82$\pm$3.17 & 0.012 && 76.11$\pm$7.36 & 0.027 && 76.77$\pm$5.30 & 0.031 && \textbf{76.79$\pm$3.97} & 0.049 \\
		($306\times 3$) & ($2^{-2}, 2^{0}, 2^{-3}$) &  && ($2^{-3}, 2^{-4}, 2^{-3}$) &  && ($2^{0}, 2^{-6}, 11$) &  && ($2^{0}, 2^{-2}, 2^{-3}, 3$) &  && ($2^{0}, 2^{2}, 2^{-2}, 3$) & \\ 
		Pima-Indian & 78.65$\pm$4.11 & 0.089 && 78.26$\pm$3.52 & 0.059 && 77.22$\pm$3.90 & 0.193 && 78.78$\pm$3.36 & 0.191 && \textbf{78.91$\pm$2.45} & 0.248 \\
		($768\times 8$) & ($2^{-2}, 2^{-2}, 2^{-2}$) &  && ($2^{-1}, 2^{-6}, 2^{-2}$) &  && ($2^{2}, 2^{-3}, 10$) &  && ($2^{1}, 2^{-3}, 2^{-1}, 4$) &  && ($2^{2}, 2^{-2}, 2^{-1}, 7$) & \\ 
		Fertility & 88.00$\pm$8.12 & 0.003 && 89.00$\pm$10.68 & 0.002 && 88.00$\pm$6.78 & 0.005 && 90.00$\pm$7.07 & 0.005 && \textbf{91.00$\pm$3.74} & 0.017 \\
		($100\times 9$) & ($2^{-8}, 2^{-3}, 2^{-2}$) &  && ($2^{-8}, 2^{2}, 2^{1}$) &  && ($2^{-5}, 2^{1}, 2$) &  && ($2^{-3}, 2^{-1}, 2^{1}, 2$) &  && ($2^{-8}, 2^{-3}, 2^{-1}, 3$) & \\ 
		Votes & 96.55$\pm$2.41 & 0.047 && \textbf{97.01$\pm$2.00} & 0.021 && 96.55$\pm$2.91 & 0.042 && \textbf{97.01$\pm$1.38} & 0.040 && \textbf{97.01$\pm$1.56} & 0.092 \\
		($435\times 16$) & ($2^{-5}, 2^{-1}, 2^{-8}$) &  && ($2^{1}, 2^{-2}, 2^{-7}$) &  && ($2^{1}, 2^{-10}, 15$) &  && ($2^{2}, 2^{0}, 2^{-7}, 10$) &  && ($2^{2}, 2^{-5}, 2^{-9}, 11$) & \\
		&  &  && &  && &  &&  &  && & \\ 
		\textbf{Win/draw/loss} &  &  && &  && &  &&  &  && & \\ 
		\tiny RKNN-TSVM(LDMDBA) & 10/0/1  &  && 9/1/1 &  && 10/1/0 &  && 6/3/2 &  && & \\
		\midrule
		Mean accuracy & \multicolumn{2}{c}{85.41} && \multicolumn{2}{c}{85.75} && \multicolumn{2}{c}{85.10} && \multicolumn{2}{c}{86.39} && \multicolumn{2}{c}{\textbf{86.51}} \\
		\bottomrule
	\end{tabular}
	
	\label{tab:3}
\end{sidewaystable*}

From the perspective of classification accuracy, our proposed RKNN-TSVM outperforms other classifiers, i.e. TSVM and WLTSVM on most datasets. This is due to the characteristics of our RKNN-TSVM which are explained as follows:

\begin{enumerate}
	\item The proposed method gives weight to each sample with respect to the distance from its nearest neighbors. As a result, noisy samples and outliers are ignored in the production of the output model. This improved the prediction accuracy of our RKNN-TSVM. On the other hand, WLTSVM gives weight to each sample by only counting the numbers of its nearest neighbors. This approach still ignores noise and outliers. However, it is not as effective as the proposed method.
	
	\item Similar to TBSVM \cite{Shao2011}, an extra stabilizer term was added to the optimization problems of our RKNN-TSVM. Therefore, two additional parameters $c_{2}$ and $c_{3}$ in RKNN-TSVM  can be adjusted which improves the classification accuracy significantly. However, these parameters are a small fixed positive scalar in TSVM and WLTSVM.
\end{enumerate}

From Table \ref{tab:3}, it can be seen that not only our RKNN-TSVM with LDMDBA algorithm outperforms TSVM, TBSVM, and WLTSVM but also it has better prediction accuracy than the RKNN-TSVM with FSA algorithm. This further validates that using a different KNN method such as the LDMDBA algorithm may improve the classification performance of our RKNN-TSVM. 

From the training time comparison of the classifiers, TSVM is faster than WLTSVM and RKNN-TSVM. Because the major computation in TSVM involves solving two smaller-sized QPPs. However, the proposed method and WLTSVM have to find KNNs for all the training samples as well as solving two smaller-sized QPPs. In order to reduce the overall computational cost, the LDMDBA algorithm was employed. Section \ref{sec:24} investigates the effectiveness of RKNN-TSVM with LDMDBA algorithm for large scale datasets.

\begin{figure}[!t]
\centering
\includegraphics[width=0.5\columnwidth]{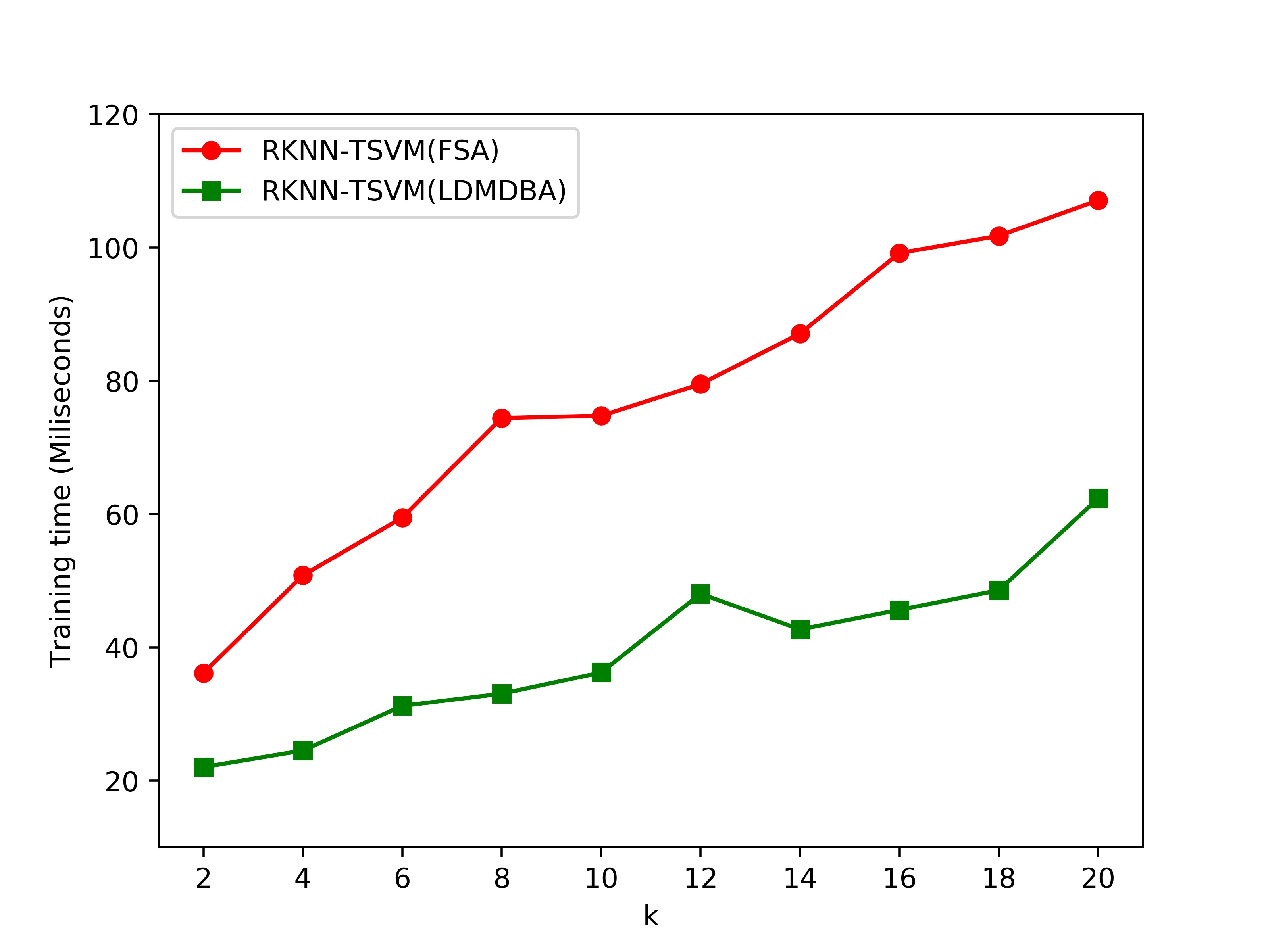}
\caption{The influence of $k$ on training time of RKNN-TSVM with FSA and LDMDBA algorithm on Pima-Indian dataset.}
\label{fig:7}
\end{figure}

Fig. \ref{fig:7} shows the influence of $k$ on training time of RKNN-TSVM with FSA and LDMDBA algorithm on Pima-Indian dataset. As shown in Fig. \ref{fig:7}, the training time of RKNN-TSVM increases with the growth of $k$. However, RKNN-TSVM with LDMDBA algorithm is significantly faster than RKNN-TSVM with FSA algorithm for each value of $k$. This also confirms our claim that the LDMDBA algorithm reduces the computational cost of the proposed method significantly.

\subsubsection{Statistical tests}\label{sec:23}
Since differences in accuracy between classifiers are not very large, non-parametric statistical tests can be used to investigate whether significant differences exist among classifiers. Hence we use Friedman test with corresponding post-hoc tests as it was suggested in Demsar \cite{demvsar2006}. This test is proved to be simple, non-parametric and safe. To run the test, average ranks of five algorithms on accuracy for all datasets were calculated and listed in Table \ref{tab:4}. Under the null-hypothesis that all the classifiers are equivalent, the Friedman test is computed according to (\ref{eq:60}):
\begin{equation}\label{eq:60}
\chi^2_F = \frac{12N}{k(k + 1)}\bigg[\sum_{j} R^2_j - \frac{k(k + 1)^2}{4} \bigg],
\end{equation}

\noindent where $R_j=\frac{1}{N}\sum_{i}r^j_i$, and $R^j_i$ denotes rank of the $j$-th of $k$ algorithms on the $i$-th of $N$ datasets. Friedman's $\chi^2_F$ is undesirably conservative and derives a better statistic
\begin{equation}\label{eq:61}
F_F = \frac{(N - 1)\chi^2_F}{N(k - 1) - \chi^2_F}
\end{equation}

\noindent which is distributed according to the $F$-distribution with $k-1$ and $(k-1)(N-1)$ degrees of freedom.

We can obtain $\chi^2_F = 24.636$ and $F_F = 12.723$ according to (\ref{eq:60}) and (\ref{eq:61}). With five classifiers and eleven datasets, $F_F$ is distributed according to $F$-distribution with $k-1$ and $(k-1)(N-1) = (4, 40)$ degrees of freedom. The critical value of $F(4,40)$ is $1.40$ for the level of significance $\alpha=0.25$, similarly, it is $2.09$ for $\alpha=0.1$ and $2.61$ for $\alpha=0.05$. Since the value of $F_F$ is much larger than the critical value, the null hypothesis is rejected. It means that there is a significant difference among five classifiers. From Table \ref{tab:4}, it can be seen that the average of RKNN-TSVM with LDMDBA algorithm is far lower than the other classifiers. 

To further analyze the performance of five classifiers statistically, we used another statistical analysis which is Win/Draw/Loss (WDL) record. The number of datasets was counted for which the proposed method with LDMDBA algorithm performs better, equally well or worse than other four classifiers. The results are shown at the end of Table \ref{tab:4}. It can be found that our RKNN-TSVM with LDMDBA algorithm is significantly better than other four classifiers.

\begin{table*}[!t]
	\small
	\centering
	\caption{Average rank on classification accuracy of five algorithms}
	\ra{1.3} 
	\begin{tabular}{c c c c c c c c c c c c c}
		\toprule
		Datasets & TSVM & TBSVM & WLTSVM & RKNN-TSVM(FSA) & RKNN-TSVM(LDMDBA) \\
		\midrule
		Australian  & 4 & 3 & 5 & 2 & 1\\ 
		Heart-Statlog  & 4 & 1.5 & 5 & 1.5 & 3\\ 
		Bupa-Liver  & 1 & 5 & 3 & 3 & 3\\ 
		WPBC  & 3 & 5 & 4 & 2 & 1\\ 
		WDBC  & 3.5 & 3.5 & 5 & 1.5 & 1.5\\ 
		Hepatitis  & 4 & 3 & 5 & 2 & 1\\ 
		Ionosphere  & 5 & 4 & 3 & 1 & 2\\ 
		Haberman  & 5 & 4 & 3 & 2 & 1\\ 
		Pima-Indian  & 3 & 4 & 5 & 2 & 1\\ 
		Fertility  & 4.5 & 3 & 4.5 & 2 & 1\\ 
		Votes  & 4.5 & 2 & 4.5 & 2 & 2\\ 
		Average rank & 3.77 & 3.45 & 4.27 & 1.91 & \textbf{1.59} \\ 
		\bottomrule
	\end{tabular}
	
	\label{tab:4}
\end{table*}

\begin{figure*}[!t]
  \centering

  \subfloat[Australian]{\includegraphics[width=0.5\columnwidth]{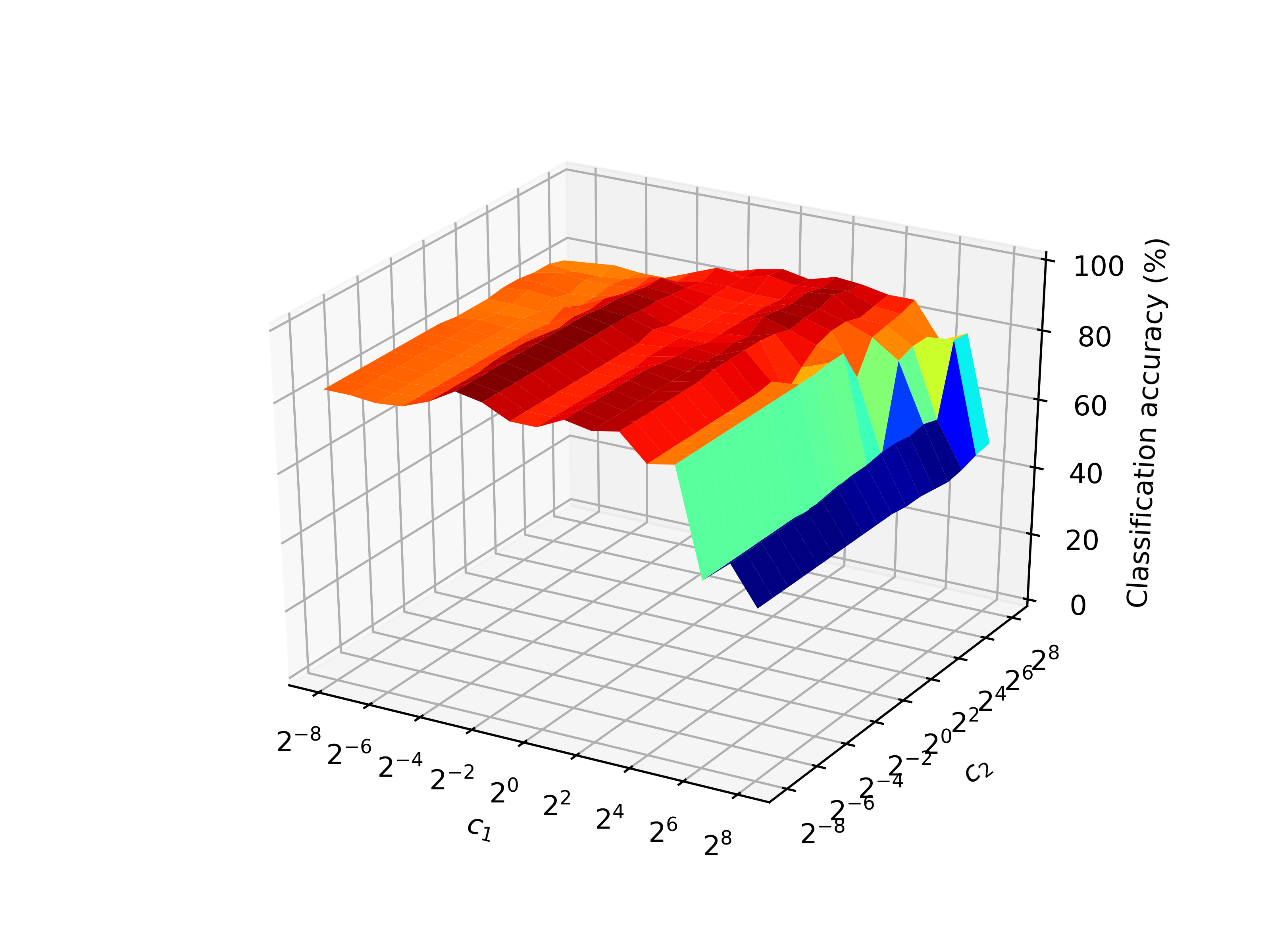}}
  \subfloat[Hepatits]{\includegraphics[width=0.5\columnwidth]{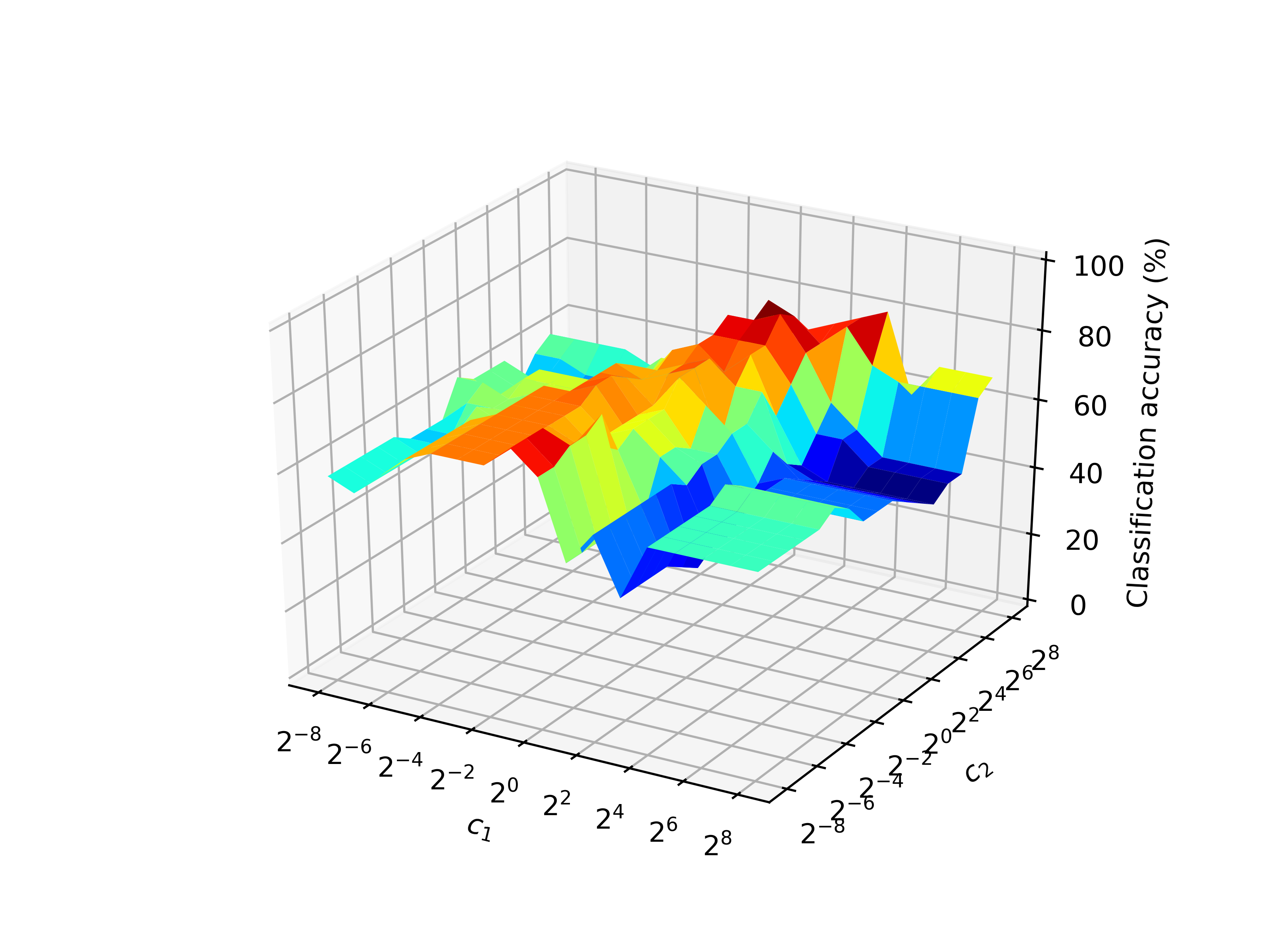}}

  \caption{The performance of linear RKNN-TSVM on parameters $c_{1}$ and $c_{2}$ for two benchmark datasets.}\label{fig:5}
\end{figure*}

\begin{figure*}[!t]
  \centering

  \subfloat[Australian]{\includegraphics[width=0.5\columnwidth]{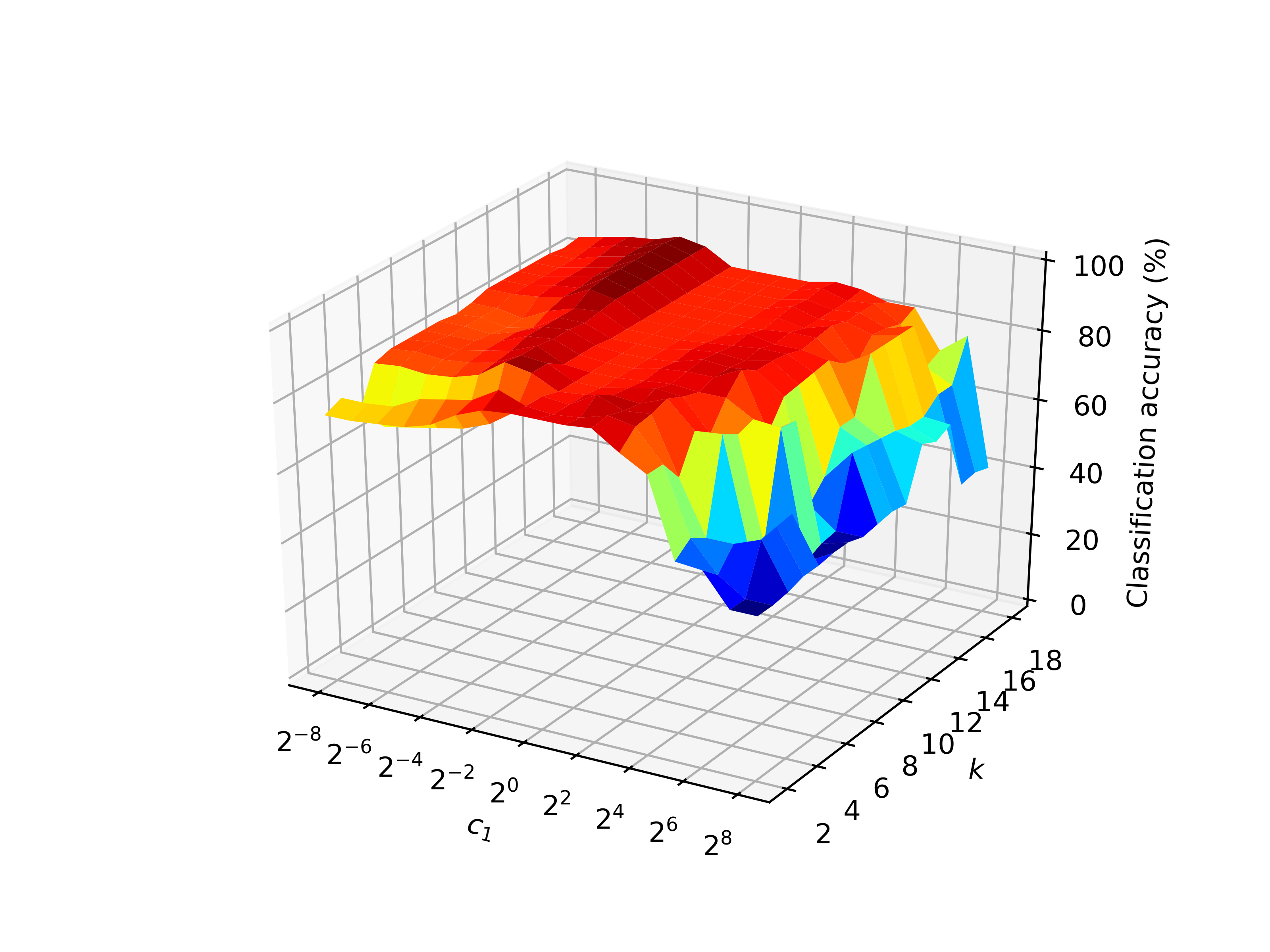}}
  \subfloat[Hepatitis]{\includegraphics[width=0.5\columnwidth]{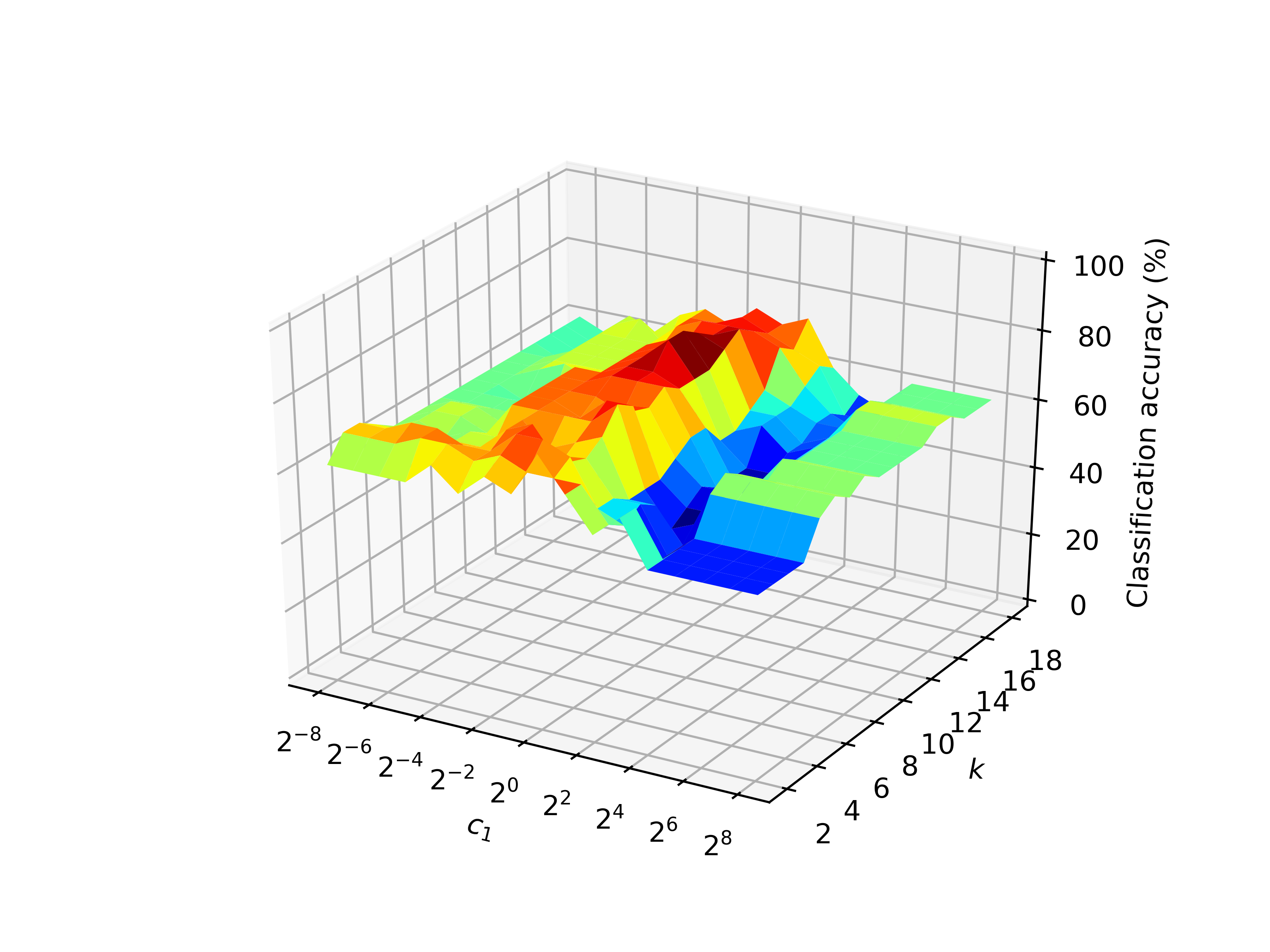}}

  \caption{The performance of linear RKNN-TSVM on parameters $c_{1}$ and $k$ for two benchmark datasets.}\label{fig:6}
\end{figure*}

\subsubsection{Parameter sensitivity}\label{sec:25}
In order to achieve a better classification accuracy, it is essential to appropriately choose parameters of our RKNN-TSVM. Hence we conducted experiments on Australian and Hepatitis datasets to analyze the sensitivity of the proposed method to parameters $c_{1}$, $c_{2}$ and $k$.

For each dataset, $c_{1}$, $c_{2}$ and $k$ can take 17 different values, resulting in 289 combinations of $(c_{1},c_{2})$ and $(c_{1}, k)$. Fig. \ref{fig:5} shows the performance of linear RKNN-TSVM on parameters $c_{1}$ and $c_{2}$ for two benchmark datasets. As can be seen from Fig. \ref{fig:5}, the values of parameter $c_{2}$ can improve the classification accuracy of the proposed method. Note that the parameter $c_{2}$ was introduced by adding a stabilizer term to objective function. This further shows that the SRM principle improves the prediction accuracy of our RKNN-TSVM. 

\indent Fig. \ref{fig:6} shows the performance of linear RKNN-TSVM on parameters $c_{1}$ and $k$ for two benchmark datasets. From Fig. \ref{fig:6}, it can be observed that the classification accuracy of our RKNN-TSVM also depends on the value of $k$. As shown in Fig \ref{fig:6}(b), the classification accuracy improves for Hepatitis dataset as the value of $k$ increases. This is because a large value of $k$ in the KNN graph reduces the effect of noisy samples and outliers on classification accuracy.

From these figures, it is clear that the prediction accuracy of RKNN-TSVM is affected by the choices of these parameters. Therefore, an appropriate selection of these parameters is crucial.

\begin{table}[!t]
  \centering
  \caption{The description of NDC datasets}
  \begin{tabular}{l c c c}
    \toprule
    Datasets & \#Training data & \#Test data & \#Features \\
    \midrule
{NDC-1K} & {1,000} & {100} & {32} \\ 
{NDC-2K} & {2,000} & {200} & {32} \\ 
{NDC-3K} & {3,000} & {300} & {32} \\ 
{NDC-4K} & {4,000} & {400} & {32} \\ 
{NDC-5K} & {5,000} & {500} & {32} \\ 
{NDC-10K} & {10,000} & {1,000} & {32} \\ 
{NDC-25K} & {25,000} & {2,500} & {32} \\ 
{NDC-50K} & {50,000} & {5,000} & {32} \\ 
    \bottomrule
  \end{tabular}
  \label{tab:5}
\end{table}

\begin{table*}[!t]
\small
  \centering
  \caption{Comparison on NDC datasets with linear kernel.}
  \ra{1.3} 
  \begin{threeparttable}
  \begin{tabular}{l c c c c c}
    \toprule
     Datasets & TSVM & WLTSVM & RKNN-TSVM(FSA) & RKNN-TSVM(LDMDBA) & \\
     & Time (s) & Time (s) & Time (s) & Time (s)  & Speedup\\
    \midrule
NDC-1K & 0.064 & 0.092 & 0.079 & 0.052 & 1.52\\ 
NDC-2K & 0.12 & 0.36 & 0.292 & 0.19 & 1.54\\ 
NDC-3K & 0.26 & 0.84 & 0.662 & 0.295 & 2.24\\ 
NDC-4K & 0.422 & 1.476 & 1.192& 0.562 & 2.12\\
    NDC-5K& 0.693 & 2.397 & 1.884& 0.828 & 2.28\\
    NDC-10K& 2.556 & 9.872 & 7.628 & 2.727  & 2.8\\
    NDC-25K& 17.606 & 68.893 & 52.867 & 16.25  & 3.25\\
    NDC-50K& 70.1 & \tnote{a} & \tnote{a} & 64.433  & -\\
    \bottomrule
  \end{tabular}
  \begin{tablenotes}
  \item[a] Experiments ran out of memory.
  \end{tablenotes}
  \end{threeparttable}
  \label{tab:6}
\end{table*}

\begin{table*}[!t]
\small
  \centering
  \caption{Comparison on NDC datasets with RBF kernel.}
  \ra{1.3} 
  \begin{threeparttable}
  \begin{tabular}{l c c c c c}
    \toprule
     Datasets & TSVM & WLTSVM & RKNN-TSVM(FSA) & RKNN-TSVM(LDMDBA) & \\
     & Time (s) & Time (s) & Time (s) & Time (s)  & Speedup\\
    \midrule
NDC-1K & 0.203 & 0.803 & 0.807 & 0.555 & 1.45\\ 
NDC-2K & 0.983 & 5.731 & 5.729 & 2.442 & 2.35\\ 
NDC-3K & 2.74 & 18.225 & 18.599 & 6.465 & 2.88\\
NDC-4K & 5.896 & 42.234 & 41.784 & 12.485 & 3.35\\ 
NDC-5K & 10.328 & 84.188 & 82.507 & 21.14 & 3.9\\
NDC-10K\textsuperscript{b} & 4.605 & 67.626 & 64.721 & 8.606 & 7.52\\
NDC-25K\textsuperscript{b} & 31.459 & 983.678 & 963.341 & 67.485  & 14.27\\
NDC-50K\textsuperscript{b} & 186.761 & \tnote{a} & \tnote{a} & 357.942  & -\\
    \bottomrule
  \end{tabular}
  \begin{tablenotes}
   \item[a] We terminated the algorithm as computing time was very high.
   \item[b] A rectangular kernel with ratio of 10\% was used.
  \end{tablenotes}
  \end{threeparttable}
  \label{tab:7}
\end{table*}

\subsubsection{Experiments on NDC datasets}\label{sec:24}
In order to analyze the computational efficiency of our RKNN-TSVM on large scale datasets, we conducted experiments on NDC datasets which were generated using David Musicant’s NDC Data Generator \cite{musicant1998ndc}. The detailed description of NDC datasets is given in Table \ref{tab:5}. For experiments with NDC datasets, the penalty parameters of all classifiers were fixed to be one (i.e. $c_{1}=1$,$c_{2}=1$,$c_{3}=1$). The Gaussian kernel with $\sigma=2^{-15}$ was used for all experiments with nonlinear kernel. The neighborhood size $k$ is also $5$ for all datasets.

Table \ref{tab:6} shows the comparison of training time for TSVM, WLTSVM and our RKNN-TSVM with a linear kernel. Similar to TSVM, TBSVM solves two smaller-sized QPPs. Therefore, training time of the TBSVM is not included. The last column shows the speedup of LDMDBA algorithm which is defined as:
\begin{align*}
\small
\textrm{Speedup} = \frac{\textrm{The training time of RKNN-TSVM(FSA)}}{\textrm{The training time of RKNN-TSVM(LDMDBA)}}
\end{align*}

From Table \ref{tab:6}, it can be seen that LDMDBA algorithm makes our RKNN-TSVM obtain much faster learning speed. It can be found that when the size of the training set increases, RKNN-TSVM with LDMDBA algorithm becomes much faster than WLTSVM and RKNN-TSVM with FSA algorithm. For instance, the proposed method with LDMDBA algorithm is $3.25$ times faster than the proposed method with FSA algorithm on NDC-25K dataset. Moreover, our linear RKNN-TSVM with LDMDBA algorithm is almost as fast as linear TSVM which is evident from Table \ref{tab:6}.

Table \ref{tab:7} shows the comparison of training time for TSVM, WLTSVM and our RKNN-TSVM with RBF kernel. The results indicate that our RKNN-TSVM with LDMDBA algorithm performed several orders of magnitude faster than WLTSVM and RKNN-TSVM with FSA algorithm. As shown in Table {\ref{tab:7}}, the largest speedup is almost 14 times. However, TSVM is almost 2 times faster than RKNN-TSVM (LDMDBA) with the reduced kernel. This is because even with the reduced kernel of dimension $(n \times \bar{n})$, RKNN-TSVM with LDMDBA algorithm still requires solving two dual QPPs as well as finding KNNs for all the samples.

The experimental results of NDC datasets with RBF kernel confirmed our claim that LDMDBA algorithm is efficient for high dimensional feature space. In summary, our RKNN-TSVM with LDMDBA algorithm is much better than WLTSVM in terms of computational time.

\section{Conclusion}\label{sec:26}
In this paper, we proposed a new classifier, i.e. an enhanced regularized K-nearest neighbor-based twin support vector machine (RKNN-TSVM). The proposed method has three clear advantages over KNN-based TSVM classifier such as WLTSVM: (1) It gives weight to each sample with respect to the distance from its nearest neighbors. This improves fitting hyperplanes with highly dense samples and makes our classifier potentially more robust to outliers. (2) Our RKNN-TSVM avoids overfitting problem by adding a stabilizer term to each primal optimization problem. Hence two parameters $c_{2}$ and $c_{3}$ were introduced which are the tradeoff between overfitting and generalization. This further improved the classification ability of our proposed method. (3) The proposed method utilizes a fast KNN method, LDMDBA algorithm. Not only this algorithm makes the learning speed of our RKNN-TSVM faster than that of WLTSVM but also improves the prediction accuracy of our proposed method.

The comprehensive experimental results on several synthetic and benchmark datasets indicate the validity and effectiveness of our proposed method. Moreover, the results on NDC datasets reveal that our RKNN-TSVM is much better than WLTSVM for handling large scale datasets. For example, the largest speed up in our RKNN-TSVM with LDMDBA algorithm reaches to 14 times. There are 4 parameters in our RKNN-TSVM which increase the computational cost of parameter selection. This limitation can be addressed in the future. The high memory consumption of the proposed method is also the main topic of future research.  

\section*{References}


\begin{thebibliography}{10}
	\expandafter\ifx\csname url\endcsname\relax
	\def\url#1{\texttt{#1}}\fi
	\expandafter\ifx\csname urlprefix\endcsname\relax\def\urlprefix{URL }\fi
	\expandafter\ifx\csname href\endcsname\relax
	\def\href#1#2{#2} \def\path#1{#1}\fi
	\bibitem{Cortes1995}
	C.~Cortes, V.~Vapnik, Support-vector networks, Machine learning 20~(3) (1995)
	273--297.
	
	\bibitem{vapnik1999overview}
	V.~N. Vapnik, An overview of statistical learning theory, IEEE transactions on
	neural networks 10~(5) (1999) 988--999.
	
	\bibitem{nasiri2009ecg}
	J.~A. Nasiri, M.~Naghibzadeh, H.~S. Yazdi, B.~Naghibzadeh, Ecg arrhythmia
	classification with support vector machines and genetic algorithm, in:
	Computer Modeling and Simulation, 2009. EMS'09. Third UKSim European
	Symposium on, IEEE, 2009, pp. 187--192.
	
	\bibitem{figuera2012}
	C.~Figuera, J.~L. Rojo-{\'A}lvarez, M.~Wilby, I.~Mora-Jim{\'e}nez, A.~J.
	Caama{\~n}o, Advanced support vector machines for 802.11 indoor location,
	Signal Processing 92~(9) (2012) 2126--2136.
	
	\bibitem{roy2016}
	A.~Roy, J.~Singha, S.~S. Devi, R.~H. Laskar, Impulse noise removal using svm
	classification based fuzzy filter from gray scale images, Signal Processing
	128 (2016) 262--273.
	
	\bibitem{wang2017}
	C.~Wang, X.~Wang, C.~Zhang, Z.~Xia, Geometric correction based color image
	watermarking using fuzzy least squares support vector machine and bessel k
	form distribution, Signal Processing 134 (2017) 197--208.
	
	\bibitem{Nayak2015}
	J.~Nayak, B.~Naik, H.~Behera, A comprehensive survey on support vector machine
	in data mining tasks: applications \& challenges, International Journal of
	Database Theory and Application 8~(1) (2015) 169--186.
	
	\bibitem{Mangasarian}
	O.~L. Mangasarian, E.~W. Wild, Proximal support vector machine classifiers, in:
	Proceedings KDD-2001: Knowledge Discovery and Data Mining, Citeseer, 2001.
	
	\bibitem{Lin2002}
	C.-F. Lin, S.-D. Wang, Fuzzy support vector machines, IEEE Transactions on
	neural networks 13~(2) (2002) 464--471.
	
	\bibitem{Mangasarian2006}
	O.~L. Mangasarian, E.~W. Wild, Multisurface proximal classification via
	generalized eigenvalues, IEEE transactions on pattern analysis and machine
	intelligence 28~(1) (2006) 69--74.
	
	\bibitem{Khemchandani2007}
	Jayadeva, R.~Khemchandani, S.~Chandra, Twin support vector machines for pattern
	classification, IEEE Transactions on pattern analysis and machine
	intelligence 29~(5).
	
	\bibitem{ding2014overview}
	S.~Ding, J.~Yu, B.~Qi, H.~Huang, An overview on twin support vector machines,
	Artificial Intelligence Review 42~(2) (2014) 245--252.
	
	\bibitem{ding2017twin}
	S.~Ding, N.~Zhang, X.~Zhang, F.~Wu, Twin support vector machine: theory,
	algorithm and applications, Neural Computing and Applications 28~(11) (2017)
	3119--3130.
	
	\bibitem{huang2018twin}
	H.~Huang, X.~Wei, Y.~Zhou, Twin support vector machines: A survey,
	Neurocomputing 300 (2018) 34--43.
	
	\bibitem{Ye2012}
	Q.~Ye, C.~Zhao, S.~Gao, H.~Zheng, Weighted twin support vector machines with
	local information and its application, Neural Networks 35 (2012) 31--39.
	
	\bibitem{Nasiri2014}
	J.~A. Nasiri, N.~M. Charkari, K.~Mozafari, Energy-based model of least squares
	twin support vector machines for human action recognition, Signal Processing
	104 (2014) 248--257.
	
	\bibitem{Pan2015}
	X.~Pan, Y.~Luo, Y.~Xu, K-nearest neighbor based structural twin support vector
	machine, Knowledge-Based Systems 88 (2015) 34--44.
	
	\bibitem{qi2013structural}
	Z.~Qi, Y.~Tian, Y.~Shi, Structural twin support vector machine for
	classification, Knowledge-Based Systems 43 (2013) 74--81.
	
	\bibitem{xu2016k}
	Y.~Xu, K-nearest neighbor-based weighted multi-class twin support vector
	machine, Neurocomputing 205 (2016) 430--438.
	
	\bibitem{xu2013twin}
	Y.~Xu, R.~Guo, L.~Wang, A twin multi-class classification support vector
	machine, Cognitive computation 5~(4) (2013) 580--588.
	
	\bibitem{pang2018scaling}
	X.~Pang, C.~Xu, Y.~Xu, Scaling knn multi-class twin support vector machine via
	safe instance reduction, Knowledge-Based Systems 148 (2018) 17--30.
	
	\bibitem{shalev2014}
	S.~Shalev-Shwartz, S.~Ben-David, Understanding machine learning: From theory to
	algorithms, Cambridge university press, 2014.
	
	\bibitem{xia2015location}
	S.~Xia, Z.~Xiong, Y.~Luo, L.~Dong, G.~Zhang, Location difference of multiple
	distances based k-nearest neighbors algorithm, Knowledge-Based Systems 90
	(2015) 99--110.
	
	\bibitem{friedman1977}
	J.~H. Friedman, J.~L. Bentley, R.~A. Finkel, An algorithm for finding best
	matches in logarithmic expected time, ACM Transactions on Mathematical
	Software (TOMS) 3~(3) (1977) 209--226.
	
	\bibitem{chen2007fast}
	Y.-S. Chen, Y.-P. Hung, T.-F. Yen, C.-S. Fuh, Fast and versatile algorithm for
	nearest neighbor search based on a lower bound tree, Pattern Recognition
	40~(2) (2007) 360--375.
	
	\bibitem{dudani1976distance}
	S.~A. Dudani, The distance-weighted k-nearest-neighbor rule, IEEE Transactions
	on Systems, Man, and Cybernetics~(4) (1976) 325--327.
	
	\bibitem{gou2012new}
	J.~Gou, L.~Du, Y.~Zhang, T.~Xiong, et~al., A new distance-weighted k-nearest
	neighbor classifier, J. Inf. Comput. Sci 9~(6) (2012) 1429--1436.
	
	\bibitem{Shao2011}
	Y.-H. Shao, C.-H. Zhang, X.-B. Wang, N.-Y. Deng, Improvements on twin support
	vector machines, IEEE transactions on neural networks 22~(6) (2011) 962--968.
	
	\bibitem{Golub2012}
	G.~H. Golub, C.~F. Van~Loan, Matrix computations, Vol.~3, JHU Press, 2012.
	
	\bibitem{sra2012optimization}
	S.~Sra, S.~Nowozin, S.~J. Wright, Optimization for machine learning, Mit Press,
	2012.
	
	\bibitem{mangasarian1999successive}
	O.~L. Mangasarian, D.~R. Musicant, Successive overrelaxation for support vector
	machines, IEEE Transactions on Neural Networks 10~(5) (1999) 1032--1037.
	
	\bibitem{Hsieh}
	C.-J. Hsieh, K.-W. Chang, C.-J. Lin, S.~S. Keerthi, S.~Sundararajan, A dual
	coordinate descent method for large-scale linear svm, in: Proceedings of the
	25th international conference on Machine learning, ACM, 2008, pp. 408--415.
	
	\bibitem{Peng2014}
	X.~Peng, D.~Chen, L.~Kong, A clipping dual coordinate descent algorithm for
	solving support vector machines, Knowledge-Based Systems 71 (2014) 266--278.
	
	\bibitem{ltsvm2019}
	A.~M. Mir, J.~A. Nasiri,
	\href{https://doi.org/10.21105/joss.01252}{Lighttwinsvm: A simple and fast
		implementation of standard twin support vector machine classifier}, Journal
	of Open Source Software 4 (2019) 1252.
	\newblock \href {http://dx.doi.org/10.21105/joss.01252}
	{\path{doi:10.21105/joss.01252}}.
	\newline\urlprefix\url{https://doi.org/10.21105/joss.01252}
	
	\bibitem{walt2011numpy}
	S.~v.~d. Walt, S.~C. Colbert, G.~Varoquaux, The numpy array: a structure for
	efficient numerical computation, Computing in Science \& Engineering 13~(2)
	(2011) 22--30.
	
	\bibitem{jones2014scipy}
	E.~Jones, T.~Oliphant, P.~Peterson, $\{$SciPy$\}$: open source scientific tools
	for $\{$Python$\}$.
	
	\bibitem{2011scikit}
	F.~Pedregosa, G.~Varoquaux, A.~Gramfort, V.~Michel, B.~Thirion, O.~Grisel,
	M.~Blondel, P.~Prettenhofer, R.~Weiss, V.~Dubourg, et~al., Scikit-learn:
	Machine learning in python, Journal of machine learning research 12~(Oct)
	(2011) 2825--2830.
	
	\bibitem{ripley2007pattern}
	B.~D. Ripley, Pattern recognition and neural networks, Cambridge university
	press, 2007.
	
	\bibitem{ho1996checkerboard}
	T.~Ho, E.~Kleinberg, Checkerboard dataset (1996).
	
	\bibitem{demvsar2006}
	J.~Dem{\v{s}}ar, Statistical comparisons of classifiers over multiple data
	sets, Journal of Machine learning research 7~(Jan) (2006) 1--30.
	
	\bibitem{musicant1998ndc}
	D.~Musicant, Ndc: normally distributed clustered datasets, Computer Sciences
	Department, University of Wisconsin, Madison.
\end{thebibliography}
\end{document}